\documentclass{article}

\usepackage{microtype}
\usepackage{graphicx}
\usepackage{subfigure}
\usepackage{booktabs} 
\usepackage{amsfonts}       
\usepackage{nicefrac}       
\usepackage{microtype}      

\usepackage{colortbl}

\usepackage{amssymb}
\usepackage{amsmath}
\usepackage{lmodern}
\usepackage{enumerate}
\usepackage{capt-of}
\usepackage{epstopdf}
\graphicspath{{fig_paper/}}
\usepackage{wrapfig}
\usepackage{multirow}
\usepackage[ruled]{algorithm2e}
\usepackage{algorithmic}

\usepackage{tabularx}
\usepackage{afterpage}
\usepackage{floatflt}
\usepackage{array}
\usepackage{arydshln}
\usepackage{makecell}
\usepackage{xcolor}
\usepackage{tikz}
\usepackage{bbm}
\usepackage{graphbox}

\usepackage[breaklinks=true]{hyperref}

\def\R{\mathbb{R}}

\def\D{\mathcal{D}}

\newcommand{\norm}[1]{\left\|#1\right\|}

\def\bydef{:=}

\def\argmax{\mathop{\rm arg\,max}\limits}

\def\maxop{\mathop{\rm max}\limits}

\def\max{\mathop{\rm max}\nolimits}

\def\Combo{{\it AutoAttack}}
\def\niter{$N_\textrm{iter}$}

\def\apgd{APGD}
\def\xorig{x_\textrm{orig}}

\newcolumntype{C}[1]{>{\centering\arraybackslash}p{#1}}
\newcolumntype{L}[1]{>{\raggedright\arraybackslash}p{#1}}
\newcolumntype{R}[1]{>{\raggedleft\arraybackslash}p{#1}}

\newcommand{\comment}[1]{}

\newcommand*{\MinNumber}{0.0}%
\newcommand*{\MidNumber}{0.5} %
\newcommand*{\MaxNumber}{1.0}%

\newcommand{\ApplyGradient}[1]{%
	\ifdim #1 pt > \MidNumber pt
	\pgfmathsetmacro{\PercentColor}{max(min(100.0*(#1 - \MidNumber)/(\MaxNumber-\MidNumber),100.0),0.00)} %
	\hspace{-0.33em}\colorbox{green!\PercentColor!yellow}{#1}
	\else
	\pgfmathsetmacro{\PercentColor}{max(min(100.0*(\MidNumber - #1)/(\MidNumber-\MinNumber),100.0),0.00)} %
	\hspace{-0.33em}\colorbox{red!\PercentColor!yellow}{#1}
	\fi
}

\newcolumntype{G}{>{\collectcell\ApplyGradient}c<{\endcollectcell}}

\def\SPSB#1#2{\rlap{\textsuperscript{\textcolor{black}{#1}}}\SB{#2}}
\def\SP#1{\textsuperscript{\textcolor{black}{#1}}}
\def\SB#1{\textsubscript{\textcolor{black}{#1}}}

\newcommand{\iter}[2]{#1^{(#2)}}
\definecolor{Gray}{gray}{0.5}

\usepackage[accepted]{icml2020}

\newif\ifpaper
\papertrue

\icmltitlerunning{Reliable Evaluation of Adversarial Robustness with an Ensemble of Diverse Parameter-free Attacks}

\begin{document}

\twocolumn[
\icmltitle{Reliable Evaluation of Adversarial Robustness with an Ensemble of Diverse Parameter-free Attacks}



\icmlsetsymbol{equal}{*}

\begin{icmlauthorlist}
\icmlauthor{Francesco Croce}{aff}
\icmlauthor{Matthias Hein}{aff}
\end{icmlauthorlist}

\icmlaffiliation{aff}{University of T{\"u}bingen, Germany}

\icmlcorrespondingauthor{F. Croce}{francesco.croce@uni-tuebingen.de}

\icmlkeywords{Machine Learning, ICML}

\vskip 0.3in
]



\printAffiliationsAndNotice{}  

\begin{abstract}
The field of defense strategies against adversarial attacks has significantly grown over the last years,
but progress is hampered as the evaluation of adversarial defenses is often insufficient
and thus gives a wrong impression of robustness. Many promising defenses could be broken later on, making it 
difficult to identify the state-of-the-art.
Frequent pitfalls in the evaluation are improper tuning of hyperparameters of the attacks, gradient obfuscation or masking.
In this paper we first propose two extensions of the PGD-attack overcoming failures due to suboptimal step size and problems of
the objective function.
We then combine our novel attacks with two complementary existing ones to form a parameter-free, computationally affordable and 
user-independent ensemble of attacks to test adversarial robustness.
We apply our ensemble to over
50 models from papers published at recent top machine learning and computer vision venues. In all except one of the cases we achieve lower robust test accuracy than reported 
in these papers, often by more than $10\%$, identifying several broken defenses.
\end{abstract}

\section{Introduction}
Adversarial samples, small perturbations of the input, with respect to some distance measure, which change the decision of a classifier,
are a problem for safe and robust machine learning. In particular, they are a major concern when it comes to safety-critical applications.
In recent years many defenses have been proposed but with more powerful or adapted attacks most of them could be broken \cite{CarWag2017, AthEtAl2018, MosEtAl18}.
Adversarial training \cite{MadEtAl2018} is one of the few approaches which could not be defeated so far. Recent variations are using other losses \cite{ZhaEtAl2019} and boost robustness via generation of additional training data \cite{CarEtAl19,AlaEtAl19} or pre-training \cite{pmlr-v97-hendrycks19a}. Another line of work 
are provable defenses, either deterministic \cite{WonEtAl18,CroEtAl2018,MirGehVec2018,GowEtAl18} or based on randomized smoothing \cite{li2018certified,lecuyer2018certified,CohenARXIV2019}. However, these are not yet competitive with the
empirical robustness of adversarial training for datasets like CIFAR-10 with large perturbations.

Due to the many broken defenses, the field is currently in a state where it is very difficult to judge the value of a new defense without an independent
test. This limits the progress as it is not clear how to distinguish bad from good ideas. A seminal work to mitigate this issue are the guidelines for
assessing adversarial defenses by \cite{CarEtAl2019}. However, as we see in our experiments, even papers trying to follow these guidelines can fail in obtaining a proper evaluation. In our opinion the reason is that at the moment there is no protocol
which works reliably and autonomously, and does not need the fine-tuning of parameters for every new defense. 
Such protocol is what we aim at in this work.

The most popular method to test adversarial robustness is the PGD (Projected Gradient Descent) attack \cite{MadEtAl2018}, as it is computationally cheap and 
performs well in many cases. However, it has been shown that even PGD can fail \cite{MosEtAl18,CroRauHei2019} leading to significant overestimation
of robustness: we identify i) the fixed step size and ii) the widely used cross-entropy loss
as two reasons for potential failure.
As remedies we propose i) a new gradient-based scheme, Auto-PGD, which does not require a step size to be chosen (Sec. \ref{sec:apgd}), and ii) an alternative loss function (Sec. \ref{sec:new_loss}).
These novel tools lead to two variants of PGD
whose only free parameter is the number of iterations, while
everything else is adjusted automatically: this is the first piece of the proposed evaluation protocol.

Another cause of poor evaluations is the lack of diversity among the attacks used, as most papers rely solely on the results given by PGD or weaker versions of it like FGSM \cite{GooShlSze2015}. Of different nature are for example two existing attacks: the white-box FAB-attack \cite{CroHei2019} and the black-box Square Attack \cite{ACFH2019square}. Importantly, these methods have a limited amount of parameters which generalize well across classifiers and datasets. In Sec. \ref{sec:combo}, we combine our two new versions of PGD with FAB and Square Attack to form a parameter-free, computationally affordable and user-independent ensemble of complementary attacks to estimate adversarial robustness, named \Combo{}.

We test \Combo{} in a large-scale evaluation (Sec. \ref{sec:exps}) on over 50 classifiers from 35 papers proposing robust models, including randomized defenses, from recent leading
conferences. Although using only five restarts for each of the three white-box attacks contained in \Combo{}, in all except two cases the robust test accuracy obtained by  \Combo{} is lower than the one reported in the original papers (our slightly more expensive \Combo{}+ is better in all but one case). For 13 models we reduce the robust accuracy by more than $10\%$ and identify several broken defenses.

We do not argue that \Combo{} is the ultimate adversarial attack
but rather that it should become the minimal test for any new defense, since it reliably reaches good performance in all tested models, \emph{without any hyperparameter tuning and at a relatively low computational cost}.
At the same time our large-scale evaluation
identifies the current state-of-the-art and which of the recent ideas are actually effective.

\section{Adversarial examples and PGD}
Let $g:\D \subseteq \R^d \longrightarrow \R^K$ be a $K$-class classifier taking decisions according to $\argmax_{k=1,\ldots, K}g_k(\cdot)$ and $\xorig\in\R^d$ a point which is correctly classified by $g$ as $c$. Given a metric $d(\cdot,\cdot)$ and $\epsilon>0$, the threat model (feasible set of the attack) is defined as $\{z \in \D  \,|\, d(\xorig,z)\leq \epsilon\}$. Then $z$ is an adversarial sample for $g$ at $\xorig$ wrt the threat model if
\[\argmax_{k=1,\ldots, K}g_k(z) \neq c, \quad d(\xorig, z)\leq \epsilon\quad \textrm{and}\quad z\in \D. \] 
To find $z$ it is common to define some surrogate function $L$ such that solving the constrained optimization problem \begin{equation} \maxop_{z\in\D} L(g(z), c) \quad \textrm{such that} \quad \gamma(\xorig, z)\leq \epsilon, \; z \in \D \label{eq:adv_ex_opt_probl} \end{equation} enforces $z$ not to be assigned to class $c$.
In image classification, the most popular threat models are based on $l_p$-distances, i.e.
$d(x,z)\bydef \norm{z-x}_p$, and $\D = [0,1]^d$. Since the projection on the $l_p$-ball for $p\in \{2, \infty\}$ is available in closed form, Problem \eqref{eq:adv_ex_opt_probl} can be solved with \textit{Projected Gradient Descent} (PGD).  
Given $f:\R^d \longrightarrow \R$, $\mathcal{S}\subset \R^d$  and the problem
$\maxop_{x \in \mathcal{S}} \; f(x)$, 
the iterations of PGD are defined for $k=1,\ldots, N_\textrm{iter}$ as
$x^{(k+1)} = P_\mathcal{S}\left(x^{(k)} + \eta^{(k)} \nabla f(x^{(k)})\right)$,
where $\eta^{(k)}$ is the step size at iteration $k$ and $P_\mathcal{S}$ is the projection onto $\mathcal{S}$. Using the cross-entropy (CE) loss as objective $L$, \cite{KurGooBen2016a, MadEtAl2018} introduced the so-called \textbf{PGD-attack}, which is currently the most popular white-box attack. In their formulation $\iter{\eta}{k} = \eta$ for every $k$, i.e. the step size is fixed, and as initial point $\iter{x}{0}$ either $\xorig$ or $\xorig + \zeta$ is used, where $\zeta$ is randomly sampled such that $\iter{x}{0}$ satisfies the constraints. Moreover, steepest
descent is done according to the norm of the threat model (e.g. for $l_\infty$ the sign of the gradient is used).

\section{Auto-PGD: A budget-aware step size-free variant of PGD}\label{sec:apgd}
We identify three weaknesses in the standard formulation of the PGD-attack and how it is used in the context of adversarial robustness. First, the \textbf{fixed step size} is suboptimal, as even for convex problems this does not guarantee convergence, and the performance of the algorithm is highly influenced by the choice of its value, see e.g. \cite{MosEtAl18}. Second, the overall scheme is in general \textbf{agnostic of the budget} given to the attack: as we show, the loss plateaus after a few iterations, except for extremely small step sizes, which however do not translate into better results. As a consequence, judging the strength of an attack by the number of iterations is misleading, see also \cite{CarEtAl2019}. Finally, the algorithm is \textbf{unaware of the trend}, i.e. does not consider whether the optimization is evolving successfully and is not able to react to this.
\begin{algorithm}[tb]
	\caption{APGD}\label{alg:apgd}
	\begin{algorithmic}[1]
		\STATE {\bfseries Input:} $f$, $S$, $x^{(0)}$, $\eta$, \niter, $W=\{w_0, \ldots, w_n\}$ %
		\STATE {\bfseries Output:} $x_\textrm{max}$, $f_\textrm{max}$
		\STATE $\iter{x}{1} \gets P_\mathcal{S}\left(x^{(0)} + \eta \nabla f(x^{(0)})\right)$
		\STATE $f_\textrm{max}\gets \max\{f(\iter{x}{0}), f(\iter{x}{1})\}$
		\STATE $x_\textrm{max} \gets \iter{x}{0}$ \textbf{if} $f_\textrm{max}\equiv f(\iter{x}{0})$ \textbf{else} $x_\textrm{max} \gets \iter{x}{1}$ 
		\FOR{$k=1$ {\bfseries to} \niter $- 1$}
		\STATE $\iter{z}{k+1} \gets P_\mathcal{S}\left(\iter{x}{k} + \eta\nabla f(\iter{x}{k})\right)$
		\STATE $\begin{aligned}[t]\iter{x}{k+1} \gets  P_\mathcal{S} &\left(\iter{x}{k} + \alpha(\iter{z}{k+1}- \iter{x}{k})\right.\\ &\left. + (1-\alpha) (\iter{x}{k} - \iter{x}{k-1})\right)\end{aligned}$
		\IF{$f(\iter{x}{k+1}) > f_\textrm{max}$}
		\STATE $x_\textrm{max} \gets \iter{x}{k+1}$ and $f_\textrm{max}\gets f(\iter{x}{k+1})$
		\ENDIF
		\IF{$k\in W$}
		\IF{Condition 1 {\bfseries or} Condition 2}
		\STATE $\eta \gets \eta / 2$ and  $\iter{x}{k+1} \gets x_\textrm{max}$
		\ENDIF
		\ENDIF
		\ENDFOR
	\end{algorithmic}
	\end{algorithm}

\subsection{Auto-PGD (\apgd{}) algorithm}
In our automatic scheme we aim at fixing these issues. The main idea is to partition the available \niter{} iterations in an initial exploration phase, where one searches the feasible set for \textit{good} initial points, and an exploitation phase, during which one tries to maximize the knowledge so far accumulated. The transition between the two phases is managed by progressively reducing the step size. In fact, a large step size allows to move quickly in $\mathcal{S}$, whereas a smaller one more eagerly maximizes the objective function locally. However, the reduction of the step size is not a priori scheduled, but rather governed by the trend of the optimization: if the value of the objective grows sufficiently fast, then the step size is most likely proper, otherwise it is reasonable to reduce it.
While the update step in \apgd{} is standard, what distinguishes our algorithm from usual PGD is the choice of the step size across iterations, which is adapted to the overall budget and to the progress of the optimization, and that, once the step size is reduced, the maximization restarts from the best point so far found. We summarize our scheme in Algorithm \ref{alg:apgd} and analyze the main features in the following.

\textbf{Gradient step:} The update of APGD follows closely the classic algorithm and only adds a momentum term. Let $\iter{\eta}{k}$ be the step size at iteration $k$, then the update step is
\begin{equation}\label{eq:gradient_step_apgd} \begin{split} \iter{z}{k+1} = P_\mathcal{S}\left(\iter{x}{k}\right. + &\left. \iter{\eta}{k} \nabla f(\iter{x}{k})\right)\\
\iter{x}{k+1} = P_\mathcal{S} \left(\iter{x}{k}\right. + & \;\alpha \cdot (\iter{z}{k+1}- \iter{x}{k})\\  + &\left.(1-\alpha) \cdot (\iter{x}{k} -\iter{x}{k-1})\right),
\end{split} 
\end{equation}
where $\alpha\in [0,1]$ (we use $\alpha=0.75$) regulates the influence of the previous update on the current one. Since in the early iterations of APGD the step size is particularly large, we want to keep a bias from the previous steps.

\textbf{Step size selection:}
We start with step size $\iter{\eta}{0}$ at iteration $0$ (we fix $\iter{\eta}{0}=2 \epsilon$), and given a budget of \niter{} iterations, we identify checkpoints $w_0=0, w_1, \ldots, w_n$ at which the algorithm decides whether it is necessary to halve the current step size. We have two conditions: 
\begin{enumerate} \item $\sum\limits_{i=w_{j-1}}^{w_{j} - 1} \mathbf{1}_{f(\iter{x}{i+1})>f(\iter{x}{i})} < \rho \cdot(w_{j} - w_{j-1})$,
\item $\iter{\eta}{w_{j-1}} \equiv \iter{\eta}{w_j}$ and $\iter{f_\textrm{max}}{w_{j-1}} \equiv \iter{f_\textrm{max}}{w_j}$,
\end{enumerate} where $\iter{f_\textrm{max}}{k}$ is the highest objective value found in the first $k$ iterations. If one of the conditions is true, then 
step size at iteration $k=w_{j}$ is halved and $\iter{\eta}{k}\bydef \iter{\eta}{w_j}/2$ for every $k=w_j + 1, \ldots, w_{j+1}$.\\
\textit{Condition 1:} counts in how many cases since the last checkpoint $w_{j-1}$ the update step has been successful in increasing $f$. If this happened for at least a fraction $\rho$ of the total update steps, then the step size is kept as the optimization is proceeding properly (we use $\rho=0.75$).\\
\textit{Condition 2:} holds true if the step size was not reduced at the last checkpoint \textit{and} there has been no improvement in the best found objective value since the last checkpoint. This prevents getting stuck in potential cycles.
%
%
%

\textbf{Restarts from the best point:}
If at a checkpoint $w_j$ the step size gets halved, then we set $\iter{x}{w_j + 1}\bydef x_\textrm{max}$, that is we restart at the point attaining the highest objective $f_{\textrm{max}}$ so far. This makes sense as reducing $\eta$ leads to a more localized search, and this should be done in a neighborhood of the current best candidate solution.

\textbf{Exploration vs exploitation:} We want the algorithm to transit gradually from exploring the whole feasible set $\mathcal{S}$ to a local optimization. This transition is regulated by progressively reducing the step size and by the choice of when to decrease it, i.e. the checkpoints $w_j$. In practice, we want to allow a relatively long initial exploration stage and then possibly update the step size more often moving toward exploitation. In fact, with smaller step sizes the improvements in the objective function are likely more frequent but also of smaller magnitude, while the importance of taking advantage of the whole input space is testified by the success of random restarts in the usual PGD-attack.
We fix the checkpoints as $w_j = \lceil p_j$\niter$\rceil\leq$ \niter, with $p_j\in[0,1]$ defined as $p_0=0$, $p_1=0.22$ and \[p_{j+1} = p_j + \max\{p_j - p_{j-1} - 0.03, 0.06\}.\]
Note that the period length $p_{j+1}-p_j$ is reduced in each step by $0.03$ but they have at least a minimum length of $0.06$.

While the proposed scheme has a few parameters which could be adjusted, we fix them to the values indicated so that the \textbf{only free variable is the budget} \niter{}.
	
\newlength{\newl}
\setlength{\newl}{0.48\columnwidth}
\begin{figure*}[t] \flushleft
	{\small
	\begin{tabular}{*{2}{C{\columnwidth}}}
		\multicolumn{1}{c}{\hspace{5mm}\cite{MadEtAl2018}} & \multicolumn{1}{c}{\cite{ZhaEtAl2019}} \\
{\scriptsize \hspace{16mm} MNIST  - $\epsilon=0.3$  \hspace{18mm} CIFAR-10  - $\epsilon=8/255$}& {\scriptsize \hspace{5mm}MNIST  - $\epsilon=0.3$\hspace{18mm} CIFAR-10  - $\epsilon=0.031$} \\
\multicolumn{2}{l}{\rotatebox[origin=c]{90}{loss} \hfill
\includegraphics[align=c, width=\newl, clip, trim=8mm 0mm 7mm 2mm]{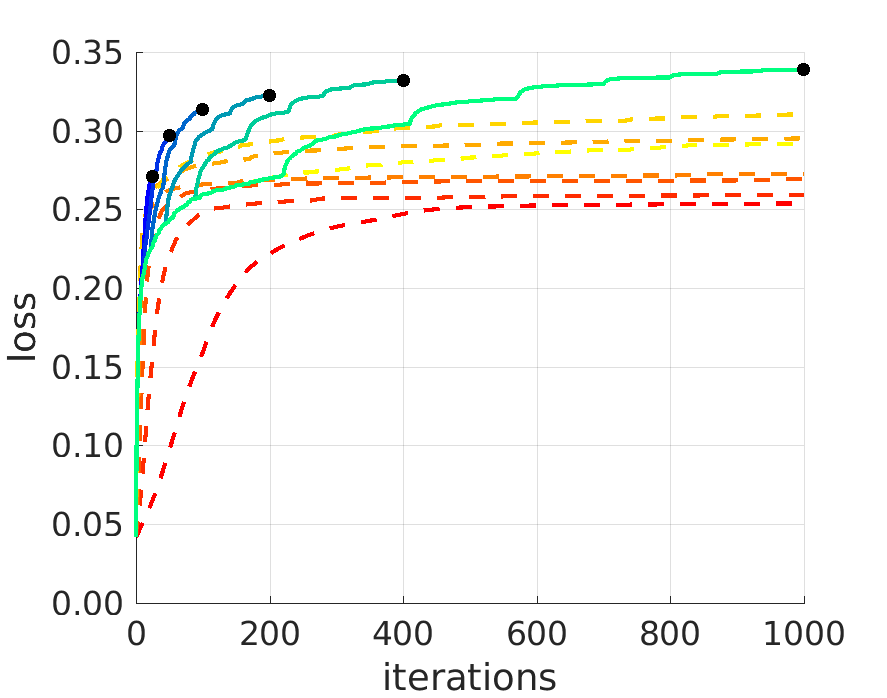} 
\includegraphics[align=c, width=\newl, clip, trim=8mm 0mm 7mm 2mm]{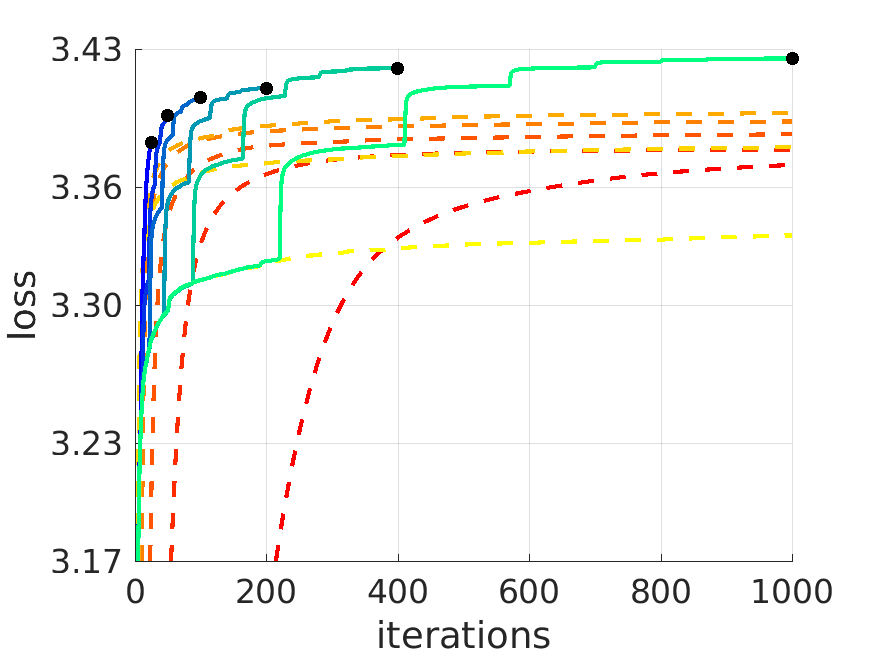}
\includegraphics[align=c, width=\newl, clip, trim=8mm 0mm 7mm 2mm]{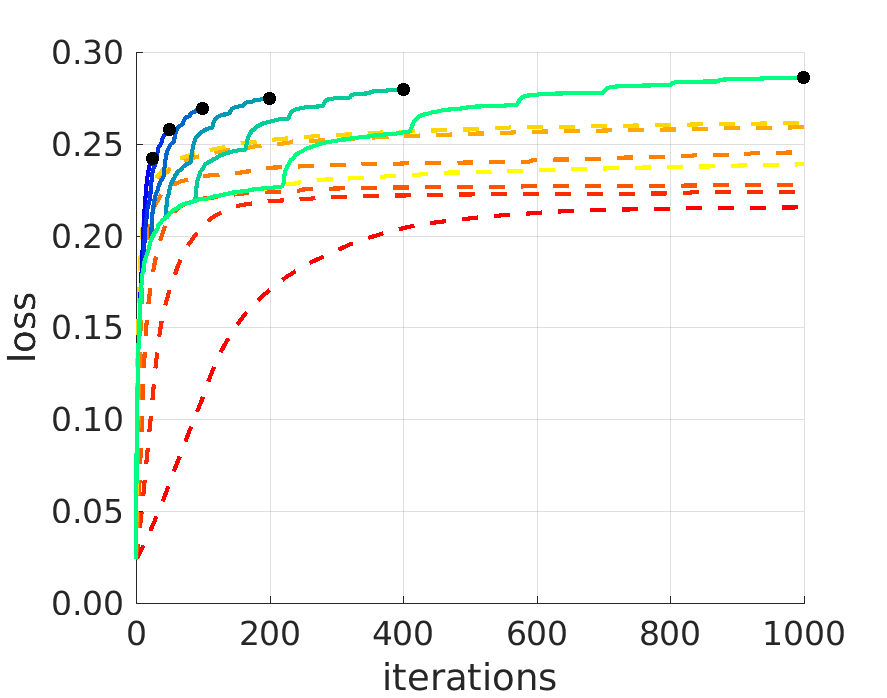} 
\includegraphics[align=c, width=\newl, clip, trim=8mm 0mm 7mm 2mm]{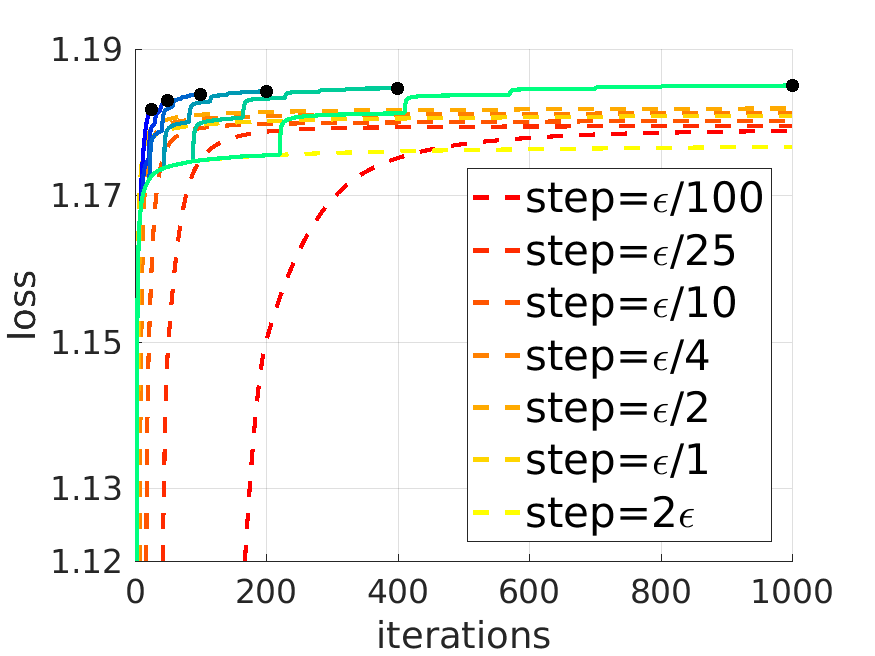}}\\

\multicolumn{2}{l}{\rotatebox[origin=c]{90}{robust accuracy} \hfill
\includegraphics[align=c, width=\newl, clip, trim=8mm 0mm 7mm 2mm]{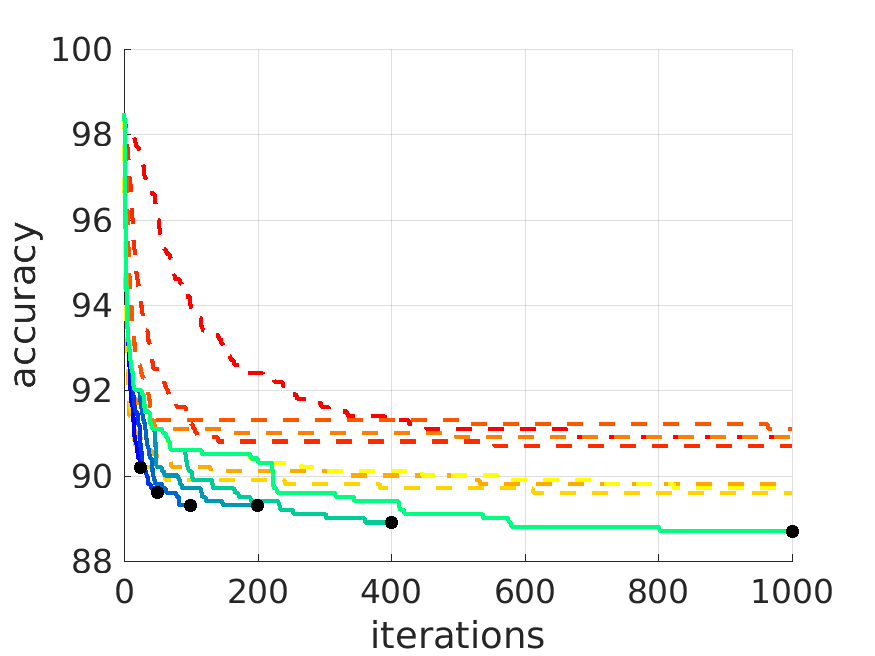}
\includegraphics[align=c, width=\newl, clip, trim=8mm 0mm 7mm 2mm]{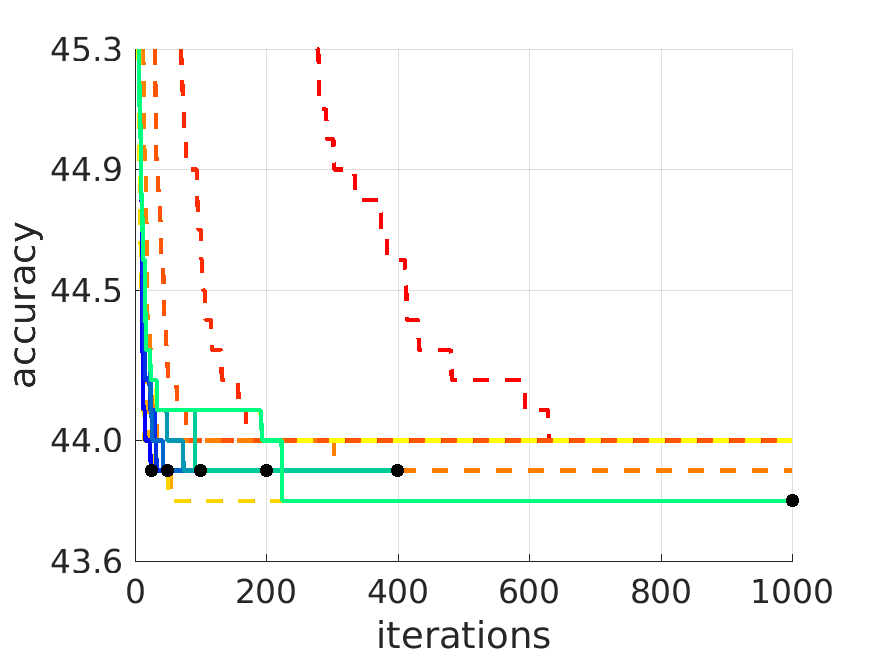}
\includegraphics[align=c, width=\newl, clip, trim=8mm 0mm 7mm 2mm]{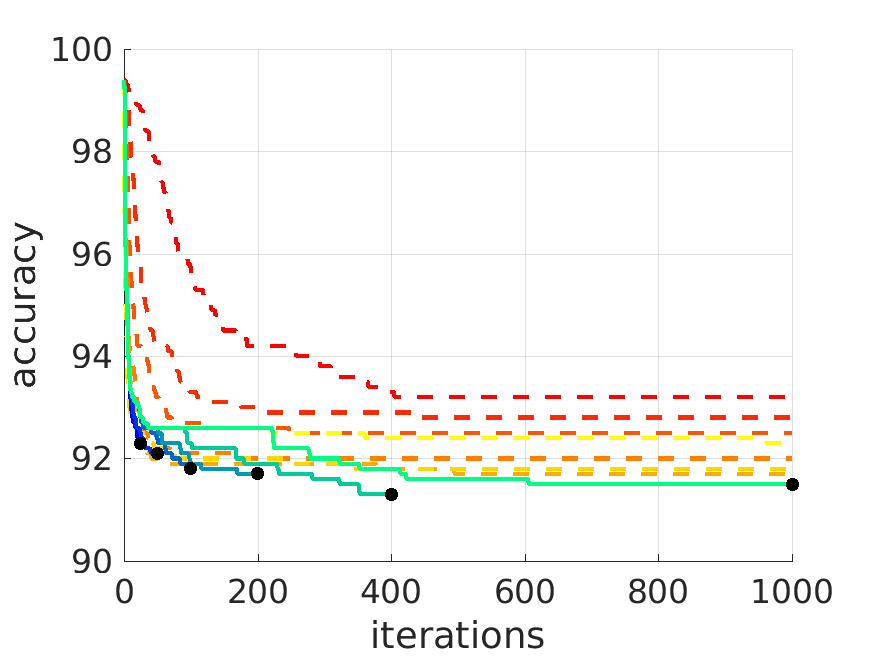}
\includegraphics[align=c, width=\newl, clip, trim=8mm 0mm 7mm 2mm]{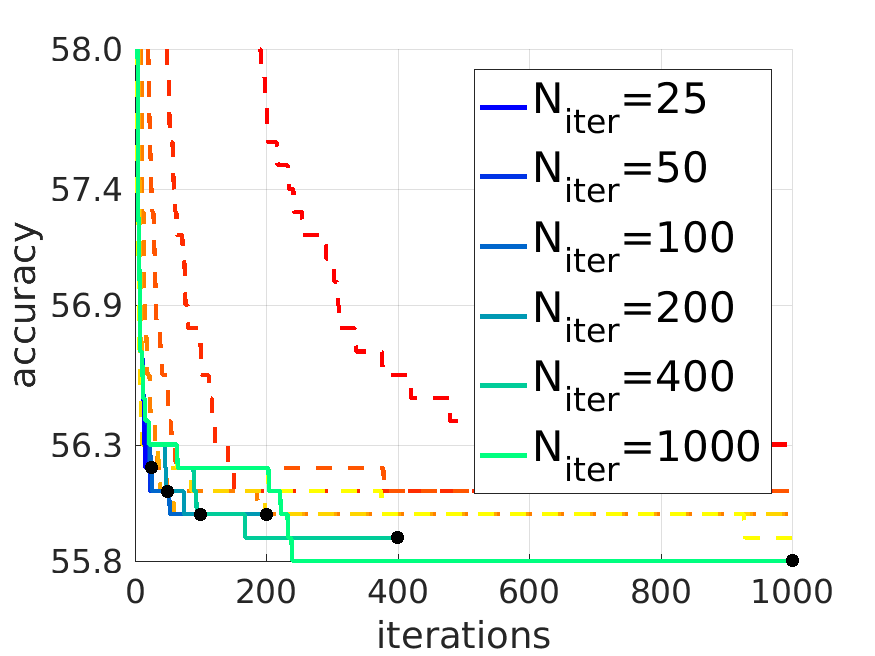}}\\
\end{tabular}}
\caption{\textbf{PGD with Momentum vs \apgd{}:} best cross-entropy loss (top) and robust accuracy (bottom) obtained so far as function of iterations for the models of \cite{MadEtAl2018} and TRADES \cite{ZhaEtAl2019} for PGD with a momentum term ($\alpha=0.75$, as used in \apgd{}) (dashed lines) with different fixed step sizes (always $1000$ iterations) and \apgd{} (solid lines) with
	different budgets of iterations. \apgd{} outperforms PGD with Momentum for every budget of iterations in achieved loss and almost always in robust accuracy.} \label{fig:pgdmomentum_vs_apgd}
\end{figure*}

\subsection{Comparison of \apgd{} to usual PGD} \label{sec:apgd_vs_pgdmomentum}
We compare our \apgd{} to PGD with Momentum in terms of achieved CE loss and robust accuracy,
focusing here on $l_\infty$-attacks with perturbation size $\epsilon$. 
We attack the robust models on MNIST and CIFAR-10 from \cite{MadEtAl2018} and \cite{ZhaEtAl2019}. We run 1000 iterations of PGD with Momentum with step sizes $\epsilon/t$  with $t\in\{0.5, 1, 2, 4, 10, 25, 100\}$, and \apgd{} with a budget of \niter{} $\in\{25, 50, 100, 200, 400, 1000\}$ iterations. 
In Figure~\ref{fig:pgdmomentum_vs_apgd} we show the evolution of the current best average cross-entropy loss and robust accuracy (i.e. the percentage of points for which the attack could not find an adversarial example) for 1000 points of the test as a function of iterations. In all cases \apgd{} achieves the highest loss (higher is better as it is a maximization problem) and this holds for any budget of iterations. Similarly, \apgd{} attains always the lowest (better) robust accuracy and thus is the stronger adversarial attack on these models (see \ifpaper Tables \ref{tab:app_pgd_vs_apgd_ce}, \ref{tab:app_pgd_vs_apgd_cw_loss} and \ref{tab:app_pgd_vs_apgd_new_loss} \else supplements \fi for a comparison across all models and different losses).
One can observe the adaptive behaviour of \apgd{}: when the budget of iterations is larger the value of the objective (the CE loss) increases more slowly, but reaches higher values in the end. This is due to the longer exploration phase, which sacrifices smaller improvements to finally get better results. In contrast, the runs of PGD with Momentum tend to plateau at suboptimal values, regardless of the choice of the step size.
An analogous comparison of \apgd{} to PGD without momentum can be found \ifpaper in Sec. \ref{sec:apgd_vs_pgd} of the Appendix\else in the supplement\fi.

%
 
%

%

%

%
%

\section{An alternative loss}
\label{sec:new_loss}
\begin{figure}[t] \centering
	\includegraphics[clip, trim=5mm 0mm 5mm 8mm, width=0.65\columnwidth]{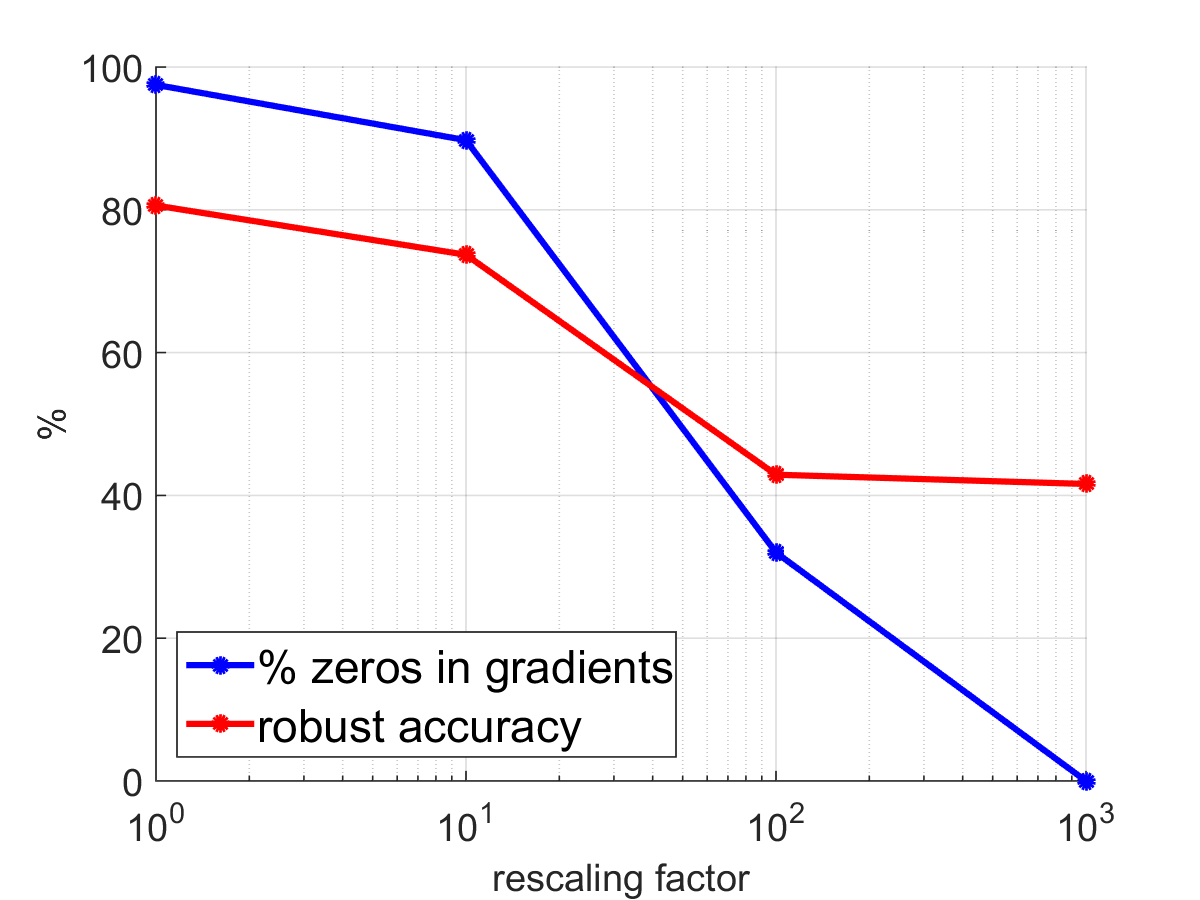} \caption{Percentage of zeros in the gradients and robust accuracy, computed by PGD on the CE loss, of the classifiers $g/\alpha$, where $g$ is the CIFAR-10 model of \cite{AtzEtAl19} and $\alpha$ a rescaling factor. The performance of PGD depends on the scale of the logits.} \label{fig:resc_factor} \end{figure}

If $x$ has correct class $y$, the cross-entropy loss at $x$ is
\begin{align} 
\textrm{CE}(x,y) = - \log p_y 
= -z_y + \log \big(\sum_{j=1}^Ke^{z_j}\big),
\end{align}
with $p_i=e^{z_i}/ \sum_{j=1}^K e^{z_j}, \; i=1,\ldots,K$, which is invariant to shifts of the logits $z$ but not to rescaling, similarly to
its gradient wrt $x$, given by
\begin{align}
\nabla_x \textrm{CE}(x,y)
=& \left(-1 + p_y\right)\nabla_x z_y + \sum_{i\neq y} p_i\nabla_x z_i. \label{eq:grad_ce}
\end{align}
If $p_y\approx 1$ and consequently $p_i\approx 0$ for $i\neq y$, then $\nabla_x \textrm{CE}(x,y)\approx \mathbf{0}$ and finite arithmetic yields $\nabla_x \textrm{CE}(x,y) = \mathbf{0}$ (this phenomenon of gradient vanishing is observed in \cite{CarWag2016}). Notice that one can achieve $p_y\approx 1$ with a classifier $h =  \alpha g$ equivalent to $g$ (i.e. they take the same decision for every $x$) but rescaled by a constant $\alpha > 0$.
To exemplify how this can lead to overestimation of robustness, we run 100 iterations of the $l_\infty$ PGD-attack on the CE loss on the CIFAR-10 model from \cite{AtzEtAl19}, with $\epsilon=0.031$, dividing the logits by a factor $\alpha \in \{1, 10^{1}, 10^2, 10^{3}\}$. In Figure \ref{fig:resc_factor} we show the fraction of entries in the gradients of $g/\alpha$ ($g$ is the original model) equal to zero and the robust accuracy achieved by the attack in dependency on $\alpha$ (we use 1000 test points, the gradient statistic is computed for correctly classified points). Without rescaling ($\alpha=1$) the gradient vanishes almost for every coordinate, so that PGD is ineffective, 
but simply rescaling the logits is sufficient to get a much more accurate robustness assessment (see also \ifpaper Sec. \ref{sec:failure}\else supplement\fi).

The CW loss \cite{CarWag2016} defined as
\begin{equation} \textrm{CW}(x,y)=-z_y + \maxop_{i\neq y} z_i. \label{eq:cw_loss}\end{equation}
has in contrast to the CE loss a direct interpretation in terms 
of the decision of the classifier. If an adversarial example exists,  then the global maximum of the CW loss is positive. However, the CW loss is not scaling invariant and thus again an extreme rescaling could in principle be used to induce gradient masking. 

\subsection{Difference of Logits Ratio Loss}
We propose the \textbf{Difference of Logits Ratio (DLR)} loss which is both shift and rescaling invariant and thus has the same degrees of freedom as the 
decision of the classifier:
\begin{equation} \textrm{DLR}(x, y) = -\frac{z_y - \maxop_{i\neq y}z_i}{z_{\pi_1} - z_{\pi_3}}, \label{eq:new_loss}\end{equation} 
where $\pi$ is the ordering of the components of $z$ in decreasing order. The required shift-invariance of the loss is achieved by having a difference of logits also
in the denominator. Maximizing DLR wrt $x$ allows to find a point classified not in class $y$ (DLR is positive only if $\mathrm{argmax}_{i}z_i \neq y$) and, once that is achieved, minimizes the score of class $y$ compared to that of the other classes.
If $x$ is correctly classified we have $\pi_1 \equiv y$, so that $\text{DLR}(x, y) = -\frac{z_{y} - z_{\pi_2}}{z_{y} - z_{\pi_3}}$ and  $\text{DLR}(x,y) \in [-1,0]$. The role of the normalization $z_{\pi_1} - z_{\pi_3}$ is to push $z_{\pi_2}$ to $z_y=z_{\pi_1}$
as it prefers points for which $z_{y} \approx z_{\pi_2} > z_{\pi_3}$ and thus is biased towards changing the decision.
Furthermore, we adapt our DLR loss to induce misclassification into a target class $t$ by \begin{equation} \textrm{Targeted-DLR}(x, y) = -\frac{z_y - z_t}{z_{\pi_1} - (z_{\pi_3} + z_{\pi_4})/2}. \label{eq:new_loss_targeted}
\end{equation} 
Thus, we preserve both the shift and scaling invariance of DLR loss, while aiming at getting $z_t > z_y$, and modify the denominator in \eqref{eq:new_loss} to ensure that the loss is not constant.

\subsection{\apgd{} versus PGD on different losses}
\label{sec:comparison_losses}
We compare 
PGD, PGD with Momentum (same momentum as in \apgd{}) and APGD with the same budget on all deterministic models trained for $l_\infty$-robustness used in the main experiments (see Sections \ref{sec:exps} and \ifpaper \ref{sec:exps_app}\else Appendix\fi) optimizing the CE, CW and DLR loss (the complete results are reported in \ifpaper Tables \ref{tab:app_pgd_vs_apgd_ce}, \ref{tab:app_pgd_vs_apgd_cw_loss}, \ref{tab:app_pgd_vs_apgd_new_loss} in Sec. \ref{sec:exps_app}\else the supplementary material\fi). For both PGD and PGD with Momentum we use three step sizes ($\epsilon/10$, $\epsilon/4$, $2\epsilon$, with $\epsilon$ the bound on the norm of the perturbations).
\apgd{} outperforms the best among the 6 versions of PGD on 32 of the 43 models with CE, 37/43 with CW and 35/43 with DLR, and the models where \apgd{} is worse are mainly the ones where the extreme step size $2\epsilon$ is optimal as the defenses lead to gradient masking/obfuscation (further details in \ifpaper Sec.~\ref{sec:bigcomparison-apgd-pgd}\else supplements\fi).
The version of standard PGD achieving most often the lowest robust accuracy is for all three losses PGD with Momentum and step size $\epsilon/4$, with average robust accuracy (over all models) of $54.84\%$, $50.42\%$ and $50.47\%$ on the CE, CW and DLR loss respectively. In the same metric, APGD achieves $54.00\%$, $49.46\%$, $48.53\%$.
Comparing CW and DLR loss per model (over all PGD versions and \apgd{}) the CW loss is up to $21\%$ worse than the DLR loss, while it is never better by more than $5\%$ and thus the DLR-loss is more stable.
In total these experiments show: i) APGD outperforms PGD/PGD with Momentum consistently regardless of the employed loss, ii) our DLR loss improves
upon the CE loss and is comparable to the CW loss, but with less severe failure cases.

We run a similar comparison for the models trained to be robust wrt $l_2$ and report the results in \ifpaper Tables \ref{tab:app_pgd_vs_apgd_ce_l2}, \ref{tab:app_pgd_vs_apgd_cw_loss_l2} and \ref{tab:app_pgd_vs_apgd_new_loss_l2}\else in the supplementary material\fi.
\apgd{} again yields most often the best results for all the losses. The difference of the losses is for $l_2$ marginal.

\section{\Combo{}: an ensemble of parameter-free attacks}\label{sec:combo}
We combine our two parameter-free versions of PGD, \apgd\textsubscript{CE} and \apgd\textsubscript{DLR}, with two existing complementary attacks, FAB \cite{CroHei2019} and Square Attack \cite{ACFH2019square}, to form the ensemble \Combo, which is automatic in the sense that it does not require to specify any free parameters\footnote{\Combo{} is available at \url{https://github.com/fra31/auto-attack}.}.

Since we want our protocol to be effective, computationally affordable and general, 
we select the following variants of the attacks to compose \Combo{}: \apgd{}\SB{CE} without random restarts, \apgd{}\SPSB{T}{DLR}, i.e. on the Targeted-DLR loss \eqref{eq:new_loss_targeted}, with 9 target classes, the targeted version of FAB, namely FAB\SP{T}, with 9 target classes and Square Attack with one run of 5000 queries. We use 100 iterations for each run of the white-box attacks.
While the runtime depends on the model, its robustness and even the framework of the target network, \apgd{} is the fastest attack, as it requires only one forward and one backward pass per iteration. The computational budget of \Combo{} is similar to what has been used, on average, in the evaluation of the defenses considered.

A key property of \Combo{} is the diversity of its components: while \apgd{} is a white-box attack aiming at any adversarial example within an $l_p$-ball, FAB minimizes the norm of the perturbation necessary to achieve a misclassification, and, although it relies on the gradient of the logits, it appears to be effective also on models affected by gradient masking as shown
in \cite{CroHei2019}. On the other hand, Square Attack is a score-based black-box attack for norm bounded perturbations
which uses random search and does not exploit any gradient approximation.
It outperforms other black-box attacks in terms of query efficiency and success rate and has been shown to be even competitive with white-box attacks \cite{ACFH2019square}. 
Both methods have few parameters which generalize well across models and datasets, so that we will keep them fixed for all experiments. The diversity within the ensemble helps for two reasons: first, there exist classifiers for which some of the attacks dramatically fail, but always at least one of them works well. Second, on the same model diverse attacks might have similar robust accuracy but succeed on different points: then considering the worst case
over all attacks, as we do for \Combo{}, improves the performance.

The hyperparameters of all attacks in \Combo{} are fixed for all experiments across datasets, models and norms \ifpaper (see Sec. \ref{sec:exp_details_app} of the Appendix)\else(see supplementary material)\fi. In Sec.~\ref{sec:exps} we show that \Combo{} evaluates adversarial robustness reliably and cost-efficiently despite its limited budget and running fully automatic, without any hyperparameter tuning.

\begin{table*}[p] \caption{\textbf{Comparison untargeted vs targeted attacks.}
		We report clean test accuracy and robust accuracy achieved by APGD\textsubscript{DLR}, \apgd{}\SPSB{T}{DLR}, FAB and FAB\SP{T}. Moreover, we show the difference between the results of the targeted and untargeted attacks, and boldface it when is negative (that is the targeted attack is stronger). 
		FAB does not scale to CIFAR-100/ImageNet due to the large number of classes.
		%
	} \label{tab:untargeted_vs_targeted}
	\vspace{2mm}
	\centering
		\small
		\begin{tabular}{r l R{10mm} ||  *{3}{R{10mm}} | *{3}{R{10mm}}} 
			\# & paper &clean&   APGD\textsubscript{DLR} &\apgd{}\SPSB{T}{DLR}& diff. &FAB & FAB\SP{T}& diff.\\
			\hline
			\multicolumn{9}{c}{}\\
			\multicolumn{9}{l}{\textbf{CIFAR-10} - $l_\infty$ - $\epsilon=8/255$}\\ \hline
			1 & \cite{CarEtAl19} & 89.69 & 60.64 & 59.54 & \textbf{-1.10} & 60.62 & 60.12 & \textbf{-0.50} \\  2 & \cite{AlaEtAl19} & 86.46 & 62.03 & 56.27 & \textbf{-5.76} & 58.20 & 56.81 & \textbf{-1.39} \\  3 & \cite{pmlr-v97-hendrycks19a} & 87.11 & 56.96 & 54.94 & \textbf{-2.02} & 55.40 & 55.27 & \textbf{-0.13} \\  4 & \cite{rice2020overfitting} & 85.34 & 55.72 & 53.43 & \textbf{-2.29} & 54.13 & 53.83 & \textbf{-0.30} \\  5 & \cite{qin2019adversarial} & 86.28 & 55.46 & 52.85 & \textbf{-2.61} & 53.77 & 53.28 & \textbf{-0.49} \\  6 & \cite{robustness} & 87.03 & 52.65 & 49.32 & \textbf{-3.33} & 50.37 & 49.81 & \textbf{-0.56} \\  7 & \cite{KumEtAl19} & 87.80 & 51.68 & 49.15 & \textbf{-2.53} & 49.87 & 49.54 & \textbf{-0.33} \\  8 & \cite{MaoEtAl19} & 86.21 & 50.33 & 47.44 & \textbf{-2.89} & 48.32 & 47.91 & \textbf{-0.41} \\  9 & \cite{ZhaEtAl19-yopo} & 87.20 & 47.33 & 44.85 & \textbf{-2.48} & 45.66 & 45.39 & \textbf{-0.27} \\  10 & \cite{MadEtAl2018} & 87.14 & 46.03 & 44.28 & \textbf{-1.75} & 45.41 & 44.75 & \textbf{-0.66} \\  11 & \cite{pang2020rethinking} & 80.89 & 44.56 & 43.50 & \textbf{-1.06} & 44.47 & 44.06 & \textbf{-0.41} \\  12 & \cite{WonEtAl20} & 83.34 & 46.64 & 43.22 & \textbf{-3.42} & 44.05 & 43.74 & \textbf{-0.31} \\  13 & \cite{ShaEtAl19} & 86.11 & 44.56 & 41.64 & \textbf{-2.92} & 42.90 & 43.44 & 0.54 \\  14 & \cite{Ding2020MMA} & 84.36 & 50.26 & 41.74 & \textbf{-8.52} & 47.18 & 42.47 & \textbf{-4.71} \\  15 & \cite{Moosavi-Dezfooli-2019-CVPR} & 83.11 & 40.29 & 38.50 & \textbf{-1.79} & 39.04 & 38.97 & \textbf{-0.07} \\  16 & \cite{ZhaWan19} & 89.98 & 48.96 & 37.29 & \textbf{-11.67} & 40.84 & 38.48 & \textbf{-2.36} \\  17 & \cite{zhang2020adversarial} & 90.25 & 49.40 & 37.54 & \textbf{-11.86} & 40.36 & 38.99 & \textbf{-1.37} \\  18 & \cite{Jang-2019-ICCV} & 78.91 & 37.01 & 34.96 & \textbf{-2.05} & 35.54 & 35.50 & \textbf{-0.04} \\  19 & \cite{kim2020sensible} & 91.51 & 48.41 & 35.93 & \textbf{-12.48} & 38.88 & 35.41 & \textbf{-3.47} \\  20 & \cite{Moosavi-Dezfooli-2019-CVPR} & 80.41 & 35.47 & 33.70 & \textbf{-1.77} & 34.08 & 34.08 & \textbf{0.00} \\  21 & \cite{Wang-2019-ICCV} & 92.80 & 40.33 & 33.61 & \textbf{-6.72} & 33.51 & 31.19 & \textbf{-2.32} \\  22 & \cite{Wang-2019-ICCV} & 92.82 & 36.58 & 29.73 & \textbf{-6.85} & 31.82 & 29.10 & \textbf{-2.72} \\  23 & \cite{Mustafa-2019-ICCV} & 89.16 & 4.54 & 1.13 & \textbf{-3.41} & 1.12 & 0.71 & \textbf{-0.41} \\  24 & \cite{pang2020rethinking} & 93.52 & 0.49 & 0.00 & \textbf{-0.49} & 0.03 & 0.00 & \textbf{-0.03} \\ \multicolumn{9}{c}{}\\ \multicolumn{9}{l}{\textbf{CIFAR-10} - $l_\infty$ - $\epsilon=0.031$} \\ \hline 1 & \cite{ZhaEtAl2019} & 84.92 & 54.04 & 53.10 & \textbf{-0.94} & 53.79 & 53.45 & \textbf{-0.34} \\  2 & \cite{AtzEtAl19} & 81.30 & 44.50 & 41.16 & \textbf{-3.34} & 40.92 & 40.73 & \textbf{-0.19} \\  3 & \cite{xiao2020enhancing} & 79.28 & 31.27 & 32.34 & 1.07 & 79.28 & 79.28 & \textbf{0.00} \\ \multicolumn{9}{c}{}\\ \multicolumn{9}{l}{\textbf{CIFAR-100} - $l_\infty$ - $\epsilon=8/255$} \\ \hline 1 & \cite{pmlr-v97-hendrycks19a} & 59.23 & 31.66 & 28.48 & \textbf{-3.18} & - & 28.74 & - \\  2 & \cite{rice2020overfitting} & 53.83 & 20.25 & 18.98 & \textbf{-1.27} & - & 19.24 & - \\ \multicolumn{9}{c}{}\\ \multicolumn{9}{l}{\textbf{MNIST} - $l_\infty$ - $\epsilon=0.3$} \\ \hline 1 & \cite{zhang2020towards} & 98.38 & 94.77 & 94.88 & 0.11 & 95.60 & 96.84 & 1.24 \\  2 & \cite{GowEtAl18} & 98.34 & 93.84 & 93.93 & 0.09 & 94.72 & 97.03 & 2.31 \\  3 & \cite{ZhaEtAl2019} & 99.48 & 93.96 & 93.58 & \textbf{-0.38} & 94.12 & 94.62 & 0.50 \\  4 & \cite{Ding2020MMA} & 98.95 & 94.03 & 94.62 & 0.59 & 94.33 & 95.37 & 1.04 \\  5 & \cite{AtzEtAl19} & 99.35 & 94.54 & 94.16 & \textbf{-0.38} & 94.12 & 95.26 & 1.14 \\  6 & \cite{MadEtAl2018} & 98.53 & 89.75 & 90.57 & 0.82 & 91.75 & 93.69 & 1.94 \\  7 & \cite{Jang-2019-ICCV} & 98.47 & 92.15 & 93.56 & 1.41 & 93.24 & 94.74 & 1.50 \\  8 & \cite{WonEtAl20} & 98.50 & 85.39 & 86.34 & 0.95 & 87.30 & 88.28 & 0.98 \\  9 & \cite{Taghanaki-2019-CVPR} & 98.86 & 0.00 & 0.00 & \textbf{0.00} & 0.02 & 0.01 & \textbf{-0.01} \\ \multicolumn{9}{c}{}\\ \multicolumn{9}{l}{\textbf{ImageNet} - $l_\infty$ - $\epsilon=4/255$} \\ \hline 1 & \cite{robustness} & 63.4 & 32.0 & 27.7 & \textbf{-4.3} & - & 28.4 & - \\ \multicolumn{9}{c}{}\\ \multicolumn{9}{l}{\textbf{CIFAR-10} - $l_2$ - $\epsilon=0.5$} \\ \hline 1 & \cite{augustin2020adversarial} & 91.08 & 74.94 & 72.91 & \textbf{-2.03} & 74.13 & 73.18 & \textbf{-0.95} \\  2 & \cite{robustness} & 90.83 & 70.20 & 69.24 & \textbf{-0.96} & 69.54 & 69.46 & \textbf{-0.08} \\  3 & \cite{rice2020overfitting} & 88.67 & 68.95 & 67.68 & \textbf{-1.27} & 68.03 & 67.97 & \textbf{-0.06} \\  4 & \cite{rony2019decoupling} & 89.05 & 67.02 & 66.44 & \textbf{-0.58} & 66.81 & 66.74 & \textbf{-0.07} \\  5 & \cite{Ding2020MMA} & 88.02 & 66.53 & 66.09 & \textbf{-0.44} & 66.43 & 66.33 & \textbf{-0.10} \\ \multicolumn{9}{c}{}\\ \multicolumn{9}{l}{\textbf{ImageNet} - $l_2$ - $\epsilon=3$} \\ \hline 1 & \cite{robustness} & 55.3 & 30.9 & 28.3 & \textbf{-2.6} & - & 28.5 & -
		\end{tabular} \end{table*}

\subsection{Untargeted vs targeted attacks} \label{sec:untargeted_vs_targeted}
We here discuss why we choose the targeted versions of \apgd{}\SB{DLR} and FAB, considering both the scalability and the effectiveness of the attacks. Note that we keep the untargeted formulation of the CE loss, as it is widely used and achieves the best results for randomized defenses (Table~\ref{tab:eval_rand_def_details}).

\textbf{FAB:} %
The untargeted version of FAB requires to compute at each iteration the Jacobian matrix of the classifier which has size scaling linearly with the number of classes $K$. While this is feasible for datasets with small $K$ (e.g. MNIST, CIFAR-10), it
becomes both computationally and memorywise prohibitive %
when many classes are given, as in the case of CIFAR-100 and ImageNet. As a solution we propose to use the targeted version of FAB, namely FAB\SP{T}, as presented in \cite{CroHei2019}, which considers only the linearization of the decision boundary between the target and the correct class, instead of all $K-1$ possible hyperplanes as for untargeted attacks, and thus by fixing additionally the number of target classes its computational complexity and memory requirements is
independent of $K$.

\textbf{Experiments:} In Table~\ref{tab:untargeted_vs_targeted} we compare the untargeted attacks, APGD\textsubscript{DLR} and FAB, to their respective targeted versions 
in terms of the robust accuracy achieved on classifiers trained with recently proposed adversarial defenses (see below for details), for different threat models. We use the untargeted methods with 5 random restarts, the targeted ones with 9 target classes, those attaining the 9 highest scores at the original point (excluding the correct one). 
One can see that on CIFAR-10, CIFAR-100 and ImageNet in almost all the cases (35/36 for \apgd{}\SB{DLR}, 29/32 for FAB) the targeted attacks provide lower (better) results, sometimes with a large gap especially for APGD\SPSB{T}{DLR}.
Although on MNIST the opposite situation occurs, we show in the next section that Square Attack is a stronger adversary than both APGD\SPSB{T}{DLR} and FAB\SP{T} for this dataset, and thus we favor the targeted attacks when selecting our ensemble in \Combo{}. %

\section{Experiments}\label{sec:exps}
In order to test the performance of \Combo, but also the individual performances of \apgd\textsubscript{CE} and \apgd\textsubscript{DLR}, %
we evaluate the adversarial robustness in the $l_\infty$- and $l_2$-threat models of over 50 models of 35 defenses from recent conferences like ICML, NeurIPS, ICLR, ICCV, CVPR, using MNIST, CIFAR-10, CIFAR-100 and ImageNet
as datasets. %
We report first results for deterministic defenses (additional ones with other thresholds $\epsilon$ in \ifpaper Sec.~\ref{sec:exps_app}\else the supplements\fi) and then for randomized ones, i.e. classifiers with a stochastic component.
\Combo{} improves almost all evaluations, showing that its good performance generalizes well across datasets, models and threat models with the same hyperparameters.

\begin{table*}[p] \caption{\textbf{Robustness evaluation of adversarial defenses by \Combo{}.}
We report clean test accuracy, the robust accuracy of the individual attacks as well as the combined one of \Combo{} (AA column). We also provide the robust accuracy reported 
in the original papers and compute the difference to the one of \Combo{}. If negative (in red) \Combo{} provides lower (better) robust accuracy. 
} \label{tab:eval_def_details}
	\vspace{2mm}
	{\centering
			\small
			\begin{tabular}{r l R{10mm} || R{10mm} *{3}{R{10mm}} >{\columncolor[rgb]{0.9 1.0 0.9}}R{10mm}| *{2}{R{10mm}}}
				\# & paper &clean&  APGD\textsubscript{CE}&APGD\SPSB{T}{DLR} & FAB\SP{T} & Square & AA & reported & reduct.\\
				\hline
				\multicolumn{10}{c}{}\\
				\multicolumn{10}{l}{\textbf{CIFAR-10} - $l_\infty$ - $\epsilon=8/255$}\\ \hline
				1 & \cite{CarEtAl19} & 89.69 &\textcolor{Gray}{\underline{61.74}} &\textcolor{Gray}{\underline{\textbf{59.54}}} &\textcolor{Gray}{\underline{60.12}} &\textcolor{Gray}{66.63} & 59.53 & 62.5 & \textcolor{red}{-2.97} \\   2 & \cite{AlaEtAl19} & 86.46 &\textcolor{Gray}{60.17} &\textcolor{Gray}{\underline{\textbf{56.27}}} &\textcolor{Gray}{56.81} &\textcolor{Gray}{66.37} & 56.03 & 56.30 & \textcolor{red}{-0.27} \\   3 & \cite{pmlr-v97-hendrycks19a} & 87.11 &\textcolor{Gray}{\underline{57.23}} &\textcolor{Gray}{\underline{\textbf{54.94}}} &\textcolor{Gray}{\underline{55.27}} &\textcolor{Gray}{61.99} & 54.92 & 57.4 & \textcolor{red}{-2.48} \\   4 & \cite{rice2020overfitting} & 85.34 &\textcolor{Gray}{\underline{57.00}} &\textcolor{Gray}{\underline{\textbf{53.43}}} &\textcolor{Gray}{\underline{53.83}} &\textcolor{Gray}{61.37} & 53.42 & 58 & \textcolor{red}{-4.58} \\   5 & \cite{qin2019adversarial} & 86.28 &\textcolor{Gray}{55.70} &\textcolor{Gray}{52.85} &\textcolor{Gray}{53.28} &\textcolor{Gray}{60.01} & 52.84 & 52.81 & \textcolor{blue}{0.03} \\   6 & \cite{robustness} & 87.03 &\textcolor{Gray}{\underline{51.72}} &\textcolor{Gray}{\underline{\textbf{49.32}}} &\textcolor{Gray}{\underline{49.81}} &\textcolor{Gray}{58.12} & 49.25 & 53.29 & \textcolor{red}{-4.04} \\   7 & \cite{KumEtAl19} & 87.80 &\textcolor{Gray}{\underline{51.80}} &\textcolor{Gray}{\underline{\textbf{49.15}}} &\textcolor{Gray}{\underline{49.54}} &\textcolor{Gray}{58.20} & 49.12 & 53.04 & \textcolor{red}{-3.92} \\   8 & \cite{MaoEtAl19} & 86.21 &\textcolor{Gray}{\underline{49.65}} &\textcolor{Gray}{\underline{\textbf{47.44}}} &\textcolor{Gray}{\underline{47.91}} &\textcolor{Gray}{56.98} & 47.41 & 50.03 & \textcolor{red}{-2.62} \\   9 & \cite{ZhaEtAl19-yopo} & 87.20 &\textcolor{Gray}{\underline{46.15}} &\textcolor{Gray}{\underline{\textbf{44.85}}} &\textcolor{Gray}{\underline{45.39}} &\textcolor{Gray}{55.08} & 44.83 & 47.98 & \textcolor{red}{-3.15} \\   10 & \cite{MadEtAl2018} & 87.14 &\textcolor{Gray}{\underline{44.75}} &\textcolor{Gray}{\underline{\textbf{44.28}}} &\textcolor{Gray}{\underline{44.75}} &\textcolor{Gray}{53.10} & 44.04 & 47.04 & \textcolor{red}{-3.00} \\   11 & \cite{pang2020rethinking} & 80.89 &\textcolor{Gray}{57.07} &\textcolor{Gray}{\underline{\textbf{43.50}}} &\textcolor{Gray}{\underline{44.06}} &\textcolor{Gray}{\underline{49.73}} & 43.48 & 55.0 & \textcolor{red}{-11.52} \\   12 & \cite{WonEtAl20} & 83.34 &\textcolor{Gray}{\underline{45.90}} &\textcolor{Gray}{\underline{\textbf{43.22}}} &\textcolor{Gray}{\underline{43.74}} &\textcolor{Gray}{53.32} & 43.21 & 46.06 & \textcolor{red}{-2.85} \\   13 & \cite{ShaEtAl19} & 86.11 &\textcolor{Gray}{\underline{43.66}} &\textcolor{Gray}{\underline{\textbf{41.64}}} &\textcolor{Gray}{\underline{43.44}} &\textcolor{Gray}{51.95} & 41.47 & 46.19 & \textcolor{red}{-4.72} \\   14 & \cite{Ding2020MMA} & 84.36 &\textcolor{Gray}{50.12} &\textcolor{Gray}{\underline{\textbf{41.74}}} &\textcolor{Gray}{\underline{42.47}} &\textcolor{Gray}{55.53} & 41.44 & 47.18 & \textcolor{red}{-5.74} \\   15 & \cite{Moosavi-Dezfooli-2019-CVPR} & 83.11 &\textcolor{Gray}{41.72} &\textcolor{Gray}{\underline{\textbf{38.50}}} &\textcolor{Gray}{\underline{38.97}} &\textcolor{Gray}{47.69} & 38.50 & 41.4 & \textcolor{red}{-2.90} \\   16 & \cite{ZhaWan19} & 89.98 &\textcolor{Gray}{64.42} &\textcolor{Gray}{\underline{\textbf{37.29}}} &\textcolor{Gray}{\underline{38.48}} &\textcolor{Gray}{\underline{59.12}} & 36.64 & 60.6 & \textcolor{red}{-23.96} \\   17 & \cite{zhang2020adversarial} & 90.25 &\textcolor{Gray}{71.40} &\textcolor{Gray}{\underline{\textbf{37.54}}} &\textcolor{Gray}{\underline{38.99}} &\textcolor{Gray}{\underline{66.88}} & 36.45 & 68.7 & \textcolor{red}{-32.25} \\   18 & \cite{Jang-2019-ICCV} & 78.91 &\textcolor{Gray}{37.76} &\textcolor{Gray}{\underline{\textbf{34.96}}} &\textcolor{Gray}{\underline{35.50}} &\textcolor{Gray}{44.33} & 34.95 & 37.40 & \textcolor{red}{-2.45} \\   19 & \cite{kim2020sensible} & 91.51 &\textcolor{Gray}{\underline{56.64}} &\textcolor{Gray}{\underline{35.93}} &\textcolor{Gray}{\underline{\textbf{35.41}}} &\textcolor{Gray}{61.30} & 34.22 & 57.23 & \textcolor{red}{-23.01} \\   20 & \cite{Moosavi-Dezfooli-2019-CVPR} & 80.41 &\textcolor{Gray}{36.65} &\textcolor{Gray}{\underline{\textbf{33.70}}} &\textcolor{Gray}{\underline{34.08}} &\textcolor{Gray}{43.46} & 33.70 & 36.3 & \textcolor{red}{-2.60} \\   21 & \cite{Wang-2019-ICCV} & 92.80 &\textcolor{Gray}{59.09} &\textcolor{Gray}{\underline{33.61}} &\textcolor{Gray}{\underline{\textbf{31.19}}} &\textcolor{Gray}{64.22} & 29.35 & 58.6 & \textcolor{red}{-29.25} \\   22 & \cite{Wang-2019-ICCV} & 92.82 &\textcolor{Gray}{69.62} &\textcolor{Gray}{\underline{29.73}} &\textcolor{Gray}{\underline{\textbf{29.10}}} &\textcolor{Gray}{\underline{66.77}} & 26.93 & 66.9 & \textcolor{red}{-39.97} \\   23 & \cite{Mustafa-2019-ICCV} & 89.16 &\textcolor{Gray}{\underline{8.16}} &\textcolor{Gray}{\underline{1.13}} &\textcolor{Gray}{\underline{\textbf{0.71}}} &\textcolor{Gray}{33.91} & 0.28 & 32.32 & \textcolor{red}{-32.04} \\   24 & \cite{chan2020jacobian} & 93.79 &\textcolor{Gray}{\underline{2.06}} &\textcolor{Gray}{\underline{\textbf{0.53}}} &\textcolor{Gray}{58.13} &\textcolor{Gray}{71.43} & 0.26 & 15.5 & \textcolor{red}{-15.24} \\   25 & \cite{pang2020rethinking} & 93.52 &\textcolor{Gray}{89.48} &\textcolor{Gray}{\underline{\textbf{0.00}}} &\textcolor{Gray}{\underline{\textbf{0.00}}} &\textcolor{Gray}{35.82} & 0.00 & 31.4 & \textcolor{red}{-31.40} \\ \multicolumn{9}{c}{}\\ \multicolumn{9}{l}{\textbf{CIFAR-10} - $l_\infty$ - $\epsilon=0.031$} \\ \hline 1 & \cite{ZhaEtAl2019} & 84.92 &\textcolor{Gray}{\underline{55.28}} &\textcolor{Gray}{\underline{\textbf{53.10}}} &\textcolor{Gray}{\underline{53.45}} &\textcolor{Gray}{59.43} & 53.08 & 56.43 & \textcolor{red}{-3.35} \\   2 & \cite{AtzEtAl19} & 81.30 &\textcolor{Gray}{79.67} &\textcolor{Gray}{\underline{41.16}} &\textcolor{Gray}{\underline{\textbf{40.73}}} &\textcolor{Gray}{47.99} & 40.22 & 43.17 & \textcolor{red}{-2.95} \\   3 & \cite{xiao2020enhancing} & 79.28 &\textcolor{Gray}{\underline{39.99}} &\textcolor{Gray}{\underline{32.34}} &\textcolor{Gray}{79.28} &\textcolor{Gray}{\underline{\textbf{20.44}}} & 18.50 & 52.4 & \textcolor{red}{-33.90} \\ \multicolumn{9}{c}{}\\ \multicolumn{9}{l}{\textbf{CIFAR-100} - $l_\infty$ - $\epsilon=8/255$} \\ \hline 1 & \cite{pmlr-v97-hendrycks19a} & 59.23 &\textcolor{Gray}{\underline{33.02}} &\textcolor{Gray}{\underline{\textbf{28.48}}} &\textcolor{Gray}{\underline{28.74}} &\textcolor{Gray}{34.26} & 28.42 & 33.5 & \textcolor{red}{-5.08} \\   2 & \cite{rice2020overfitting} & 53.83 &\textcolor{Gray}{\underline{20.57}} &\textcolor{Gray}{\underline{\textbf{18.98}}} &\textcolor{Gray}{\underline{19.24}} &\textcolor{Gray}{\underline{23.57}} & 18.95 & 28.1 & \textcolor{red}{-9.15} \\ \multicolumn{9}{c}{}\\ \multicolumn{9}{l}{\textbf{MNIST} - $l_\infty$ - $\epsilon=0.3$} \\ \hline 1 & \cite{zhang2020towards} & 98.38 &\textcolor{Gray}{\underline{95.32}} &\textcolor{Gray}{\underline{94.88}} &\textcolor{Gray}{96.84} &\textcolor{Gray}{\underline{\textbf{93.97}}} & 93.96 & 96.38 & \textcolor{red}{-2.42} \\   2 & \cite{GowEtAl18} & 98.34 &\textcolor{Gray}{94.79} &\textcolor{Gray}{93.93} &\textcolor{Gray}{97.03} &\textcolor{Gray}{\underline{\textbf{92.88}}} & 92.83 & 93.88 & \textcolor{red}{-1.05} \\   3 & \cite{ZhaEtAl2019} & 99.48 &\textcolor{Gray}{\underline{93.60}} &\textcolor{Gray}{\underline{93.58}} &\textcolor{Gray}{\underline{94.62}} &\textcolor{Gray}{\underline{\textbf{92.97}}} & 92.81 & 95.60 & \textcolor{red}{-2.79} \\   4 & \cite{Ding2020MMA} & 98.95 &\textcolor{Gray}{94.58} &\textcolor{Gray}{94.62} &\textcolor{Gray}{95.37} &\textcolor{Gray}{\underline{\textbf{91.42}}} & 91.40 & 92.59 & \textcolor{red}{-1.19} \\   5 & \cite{AtzEtAl19} & 99.35 &\textcolor{Gray}{99.10} &\textcolor{Gray}{\underline{94.16}} &\textcolor{Gray}{\underline{95.26}} &\textcolor{Gray}{\underline{\textbf{90.86}}} & 90.85 & 97.35 & \textcolor{red}{-6.50} \\   6 & \cite{MadEtAl2018} & 98.53 &\textcolor{Gray}{90.57} &\textcolor{Gray}{90.57} &\textcolor{Gray}{93.69} &\textcolor{Gray}{\underline{\textbf{88.56}}} & 88.50 & 89.62 & \textcolor{red}{-1.12} \\   7 & \cite{Jang-2019-ICCV} & 98.47 &\textcolor{Gray}{\underline{94.05}} &\textcolor{Gray}{\underline{93.56}} &\textcolor{Gray}{94.74} &\textcolor{Gray}{\underline{\textbf{88.00}}} & 87.99 & 94.61 & \textcolor{red}{-6.62} \\   8 & \cite{WonEtAl20} & 98.50 &\textcolor{Gray}{\underline{86.68}} &\textcolor{Gray}{\underline{86.34}} &\textcolor{Gray}{\underline{88.28}} &\textcolor{Gray}{\underline{\textbf{83.07}}} & 82.93 & 88.77 & \textcolor{red}{-5.84} \\   9 & \cite{Taghanaki-2019-CVPR} & 98.86 &\textcolor{Gray}{\underline{30.50}} &\textcolor{Gray}{\underline{\textbf{0.00}}} &\textcolor{Gray}{\underline{0.01}} &\textcolor{Gray}{\underline{\textbf{0.00}}} & 0.00 & 64.25 & \textcolor{red}{-64.25} \\ \multicolumn{9}{c}{}\\ \multicolumn{9}{l}{\textbf{ImageNet} - $l_\infty$ - $\epsilon=4/255$} \\ \hline 1 & \cite{robustness} & 63.4 &\textcolor{Gray}{\underline{31.0}} &\textcolor{Gray}{\underline{\textbf{27.7}}} &\textcolor{Gray}{\underline{28.4}} &\textcolor{Gray}{46.8} & 27.6 & 33.38 & \textcolor{red}{-5.78} \\ \multicolumn{9}{c}{}\\ \multicolumn{9}{l}{\textbf{CIFAR-10} - $l_2$ - $\epsilon=0.5$} \\ \hline 1 & \cite{augustin2020adversarial} & 91.08 &\textcolor{Gray}{74.70} &\textcolor{Gray}{\underline{\textbf{72.91}}} &\textcolor{Gray}{\underline{73.18}} &\textcolor{Gray}{83.10} & 72.91 & 73.27 & \textcolor{red}{-0.36} \\   2 & \cite{robustness} & 90.83 &\textcolor{Gray}{\underline{69.62}} &\textcolor{Gray}{\underline{\textbf{69.24}}} &\textcolor{Gray}{\underline{69.46}} &\textcolor{Gray}{80.92} & 69.24 & 70.11 & \textcolor{red}{-0.87} \\   3 & \cite{rice2020overfitting} & 88.67 &\textcolor{Gray}{\underline{68.58}} &\textcolor{Gray}{\underline{\textbf{67.68}}} &\textcolor{Gray}{\underline{67.97}} &\textcolor{Gray}{79.01} & 67.68 & 71.6 & \textcolor{red}{-3.92} \\   4 & \cite{rony2019decoupling} & 89.05 &\textcolor{Gray}{\underline{66.59}} &\textcolor{Gray}{\underline{\textbf{66.44}}} &\textcolor{Gray}{\underline{66.74}} &\textcolor{Gray}{78.05} & 66.44 & 67.6 & \textcolor{red}{-1.16} \\   5 & \cite{Ding2020MMA} & 88.02 &\textcolor{Gray}{66.21} &\textcolor{Gray}{\underline{\textbf{66.09}}} &\textcolor{Gray}{66.33} &\textcolor{Gray}{76.99} & 66.09 & 66.18 & \textcolor{red}{-0.09} \\ \multicolumn{9}{c}{}\\ \multicolumn{9}{l}{\textbf{ImageNet} - $l_2$ - $\epsilon=3$} \\ \hline 1 & \cite{robustness} & 55.3 &\textcolor{Gray}{\underline{31.5}} &\textcolor{Gray}{\underline{\textbf{28.3}}} &\textcolor{Gray}{\underline{28.5}} &\textcolor{Gray}{46.6} & 28.3 & 35.09 & \textcolor{red}{-6.79}
		\end{tabular}

}

\end{table*}

\begin{table*}[h!] \caption{\textbf{Robustness evaluation of randomized $l_\infty$-adversarial defenses by \Combo.} We
report the clean test accuracy (mean and standard deviation over 5 runs) and the robust accuracy of the individual attacks as well as the combined on of \Combo{} (again over 5 runs). We also provide the robust accuracy reported in the respective papers and compute the difference to the one of \Combo{} (negative means
that \Combo{} is better). The statistics of our attack are computed on the whole test set except for the ones of \cite{pmlr-v97-yang19e}, which are on 1000 test points due to the computational cost of this defense. The $\epsilon$ is the same as used in the papers.} \label{tab:eval_rand_def_details}\vspace{2mm}
	{\centering
		\setlength{\tabcolsep}{4.9pt}
		\begin{small}
		\begin{tabular}{R{1mm} L{33mm} l R{17mm} || *{4}{R{9mm}} >{\columncolor[rgb]{0.9 1.0 0.9}}R{17mm}| *{1}{R{9mm}} R{9mm}}
			\# & paper &model& clean&  
			APGD\textsubscript{CE}&APGD\textsubscript{DLR} &
			FAB & Square & \Combo{}& report. & reduct.\\
			\hline
			\multicolumn{11}{c}{}\\ \multicolumn{11}{l}{\textbf{CIFAR-10} - $\epsilon=8/255$} \\ \hline 1 & \cite{WanEtAl19} & En\textsubscript{5}RN & 82.39 (0.14)  &\textcolor{Gray}{\underline{\textbf{48.81}}} &\textcolor{Gray}{\underline{49.37}} & \textcolor{Gray}{-} & \textcolor{Gray}{78.61} & 45.56 (0.20)  & 51.48 & \textcolor{red}{-5.9} \\  2 & \cite{pmlr-v97-yang19e} & with AT & 84.9 (0.6)  &\textcolor{Gray}{\underline{\textbf{30.1}}} &\textcolor{Gray}{\underline{31.9}} & \textcolor{Gray}{-} & \textcolor{Gray}{-} & 26.3 (0.85)  & 52.8 & \textcolor{red}{-26.5} \\  3 & \cite{pmlr-v97-yang19e} & pure & 87.2 (0.3)  &\textcolor{Gray}{\underline{\textbf{21.5}}} &\textcolor{Gray}{\underline{24.3}} & \textcolor{Gray}{-} & \textcolor{Gray}{-} & 18.2 (0.82)  & 40.8 & \textcolor{red}{-22.6} \\  4 & \cite{grathwohl2020your} & JEM-10 & 90.99 (0.03)  &\textcolor{Gray}{\underline{\textbf{11.69}}} &\textcolor{Gray}{\underline{15.88}} & \textcolor{Gray}{63.07} & \textcolor{Gray}{79.32} & 9.92 (0.03)  & 47.6 & \textcolor{red}{-37.7} \\  5 & \cite{grathwohl2020your} & JEM-1 & 92.31 (0.04)  &\textcolor{Gray}{\underline{\textbf{9.15}}} &\textcolor{Gray}{\underline{13.85}} & \textcolor{Gray}{62.71} & \textcolor{Gray}{79.25} & 8.15 (0.05)  & 41.8 & \textcolor{red}{-33.6} \\  6 & \cite{grathwohl2020your} & JEM-0 & 92.82 (0.05)  &\textcolor{Gray}{\underline{\textbf{7.19}}} &\textcolor{Gray}{\underline{12.63}} & \textcolor{Gray}{66.48} & \textcolor{Gray}{73.12} & 6.36 (0.06)  & 19.8 & \textcolor{red}{-13.4} \\ \multicolumn{11}{c}{}\\ \multicolumn{11}{l}{\textbf{CIFAR-10} - $\epsilon=4/255$} \\ \hline 1 & \cite{grathwohl2020your} & JEM-10 & 91.03 (0.05)  &\textcolor{Gray}{\underline{\textbf{49.10}}} &\textcolor{Gray}{\underline{52.55}} & \textcolor{Gray}{78.87} & \textcolor{Gray}{89.32} & 47.97 (0.05)  & 72.6 & \textcolor{red}{-24.6} \\  2 & \cite{grathwohl2020your} & JEM-1 & 92.34 (0.04)  &\textcolor{Gray}{\underline{\textbf{46.08}}} &\textcolor{Gray}{\underline{49.71}} & \textcolor{Gray}{78.93} & \textcolor{Gray}{90.17} & 45.49 (0.04)  & 67.1 & \textcolor{red}{-21.6} \\  3 & \cite{grathwohl2020your} & JEM-0 & 92.82 (0.02)  &\textcolor{Gray}{\underline{\textbf{42.98}}} &\textcolor{Gray}{\underline{47.74}} & \textcolor{Gray}{82.92} & \textcolor{Gray}{89.52} & 42.55 (0.07)  & 50.8 & \textcolor{red}{-8.2}
			\end{tabular}
			\end{small}}
	\end{table*}

\textbf{Deterministic defenses:} In Tables \ref{tab:eval_def_details} \ifpaper and \ref{tab:eval_additional_defenses} (in the Appendix) \else (and in the Appendix)\fi we report the results on 49 models, 43 trained for $l_\infty$- and 6 for $l_2$-robustness, from recent defense papers (for some of them multiple networks are considered, possibly on more datasets and norms). When possible we used the originals models (which are either publicly available or we obtained them via personal communication from the authors).
Otherwise we retrained the models with the code released by the authors. Further details about the models and papers can be found in the \ifpaper Appendix (Sec. \ref{sec:exps_app})\else Appendix\fi. For each classifier we report the clean accuracy and robust accuracy, at the $\epsilon$ specified in the table, on the whole test set (except for ImageNet where we use 1000 points from the validation set) obtained by the individual attacks \apgd\textsubscript{CE}, \apgd\SPSB{T}{DLR}, FAB\SP{T} and Square Attack, together with our ensemble \Combo{}, which counts as a success every point on which \textit{at least one} of the four attacks finds an adversarial example (worst case evaluation). Additionally, we provide the \textit{reported} robust accuracy of the respective papers (please note that in some cases their statistics are computed on a subset of the test set)
and the difference between our robust accuracy and the reported one. The reduction is highlighted in red in the last column of Table \ref{tab:eval_def_details} if it is negative (we get a lower robust accuracy). Finally, we boldface the attack which obtains the best individual robust accuracy and underline those achieving a robust accuracy lower than reported.

Notably, in all but one case \Combo{} achieves a lower robust accuracy than reported in the original papers, and the improvement is larger than 10\% in 13 out of 49 cases, larger than 30\% in 8 cases (\Combo{} yields also significantly lower values on the few models on CIFAR-100 and ImageNet). Thus \Combo{} would almost always have provided  a better estimate of the robustness of the models than in the original evaluation, without any adaptation to the specific defense. In the only case where it does not reduce the reported robust accuracy it is only 0.03\% far from it, and this result has been obtained with a variant of PGD with 180 restarts and 200 iterations (see \cite{qin2019adversarial}), which is way more expensive than our evaluation. 

In most of the cases more than one of the attacks included in \Combo{} achieves a lower robust accuracy than reported
(\apgd\textsubscript{CE} improves the reported evaluation in 21/49 cases, \apgd\SPSB{T}{DLR} in 45/49, FAB\SP{T} in 39/49 and Square Attack in 17/49, but 9/9 on MNIST).
\apgd\SPSB{T}{DLR} most often attains the best result for CIFAR-10, CIFAR-100 and ImageNet, Square Attack on MNIST. Also, \apgd\SPSB{T}{DLR} is the most reliable one as it has the least severe failure which we define as the largest difference in robust accuracy to the best performing attack (maximal difference less than 12\%, compared to 89\% for \apgd\textsubscript{CE}, 59\% for FAB\SP{T} and 70\% for Square Attack). Thus our new DLR loss is able to resist gradient masking.

\textbf{Randomized defenses:} Another line of adversarial defenses relies on adding to a classifier some stochastic component. In this case the output (hence the decision) of the model might change across different runs for the same input. Thus we compute in Table \ref{tab:eval_rand_def_details} the mean (standard deviation in brackets) of our statistics over 5 runs.
Moreover the results of \Combo{} are given considering, for each point, the attack performing better on average across 5 runs.
To counter the randomness of the classifiers, for \apgd{} we compute the direction for the update step as the average of 20 computations of the gradient at the same point (known as Expectation over Transformation \cite{AthEtAl2018}) and use the untargeted losses (1 run). We do not run FAB here since it returns points on or very close to the decision boundary, so that even a small variation in the classifier is likely to undo the adversarial change. 
We modify Square Attack to accept an update if it reduces the target loss on average over 20 forward passes and, as this costs more time we use only 1000 iterations.
For the models from \cite{grathwohl2020your}, we attack that named JEM-0 with 5 restarts with the deterministic versions (i.e. without averaging across multiple passes of the networks), since the stochastic component has little influence, and then reuse the same adversarial examples on JEM-1 and JEM-10 (the results of FAB confirm that it is not suitable to test randomized defenses).
Table \ref{tab:eval_rand_def_details}
shows that \Combo{} achieves always lower robust accuracy than reported in the respective papers, with
\apgd\textsubscript{CE} being the best performing attack, closely followed by \apgd\textsubscript{DLR}.  In 7 out of 9 cases the improvement is significant, larger than 10\% (and in 3/9 cases larger than 25\%). Thus \Combo{} is also suitable for the evaluation of randomized defenses.
			
%

%

%
%

%

\subsection{Analysis of SOTA of adversarial defenses}
While the main goal of the evaluation is to show the effectiveness of \Combo{}, at the same time it provides
an assessment of the SOTA of adversarial defenses.
The most robust defenses rely on variations or fine-tuning of adversarial training introduced in \cite{MadEtAl2018}. One step forward has been made by methods which use additional data for training, like \cite{CarEtAl19} and \cite{AlaEtAl19}. Moreover, several defenses which claim SOTA robustness turn out to be significantly less robust than \cite{MadEtAl2018}.
Interestingly, the most (empirically) resistant model on MNIST is one trained for obtaining provable certificates on the exact robust accuracy, and comes with a verified lower bound on it of 93.32\% \cite{zhang2020towards}.

While this paper contains up to our knowledge the largest independent evaluation of current adversarial defenses, this is by no means an exhaustive survey.
Several authors did not reply to our request or were not able to provide models (or at least code). We thank all the authors who helped us in this evaluation. We hope that \Combo{} will contribute to a faster development of adversarial defenses and recommend it as part of a standard evaluation pipeline as it is quick and
requires no hyperparameter tuninig.

\section*{Acknowledgements}
We are very grateful to Alvin Chan, Chengzhi Mao, Seyed-Mohsen Moosavi-Dezfooli, Chongli Qin, Saeid Asgari Taghanaki, Bao Wang and Zhi Xu for providing code, models and answering questions on their papers. We also thank Maksym Andriushchenko for insightful discussions about this work. We acknowledge support from the German Federal Ministry of Education and Research (BMBF) through the T\"{u}bingen AI Center (FKZ: 01IS18039A). This work was also supported by the DFG Cluster of Excellence “Machine Learning – New Perspectives for Science”, EXC 2064/1, project number 390727645, and by DFG grant 389792660 as part of TRR~248.

%

\bibliographystyle{icml2020}


\ifpaper
\clearpage
\appendix
 
\section{Auto-PGD}
In the main paper we compare in Sec. \ref{sec:apgd_vs_pgdmomentum} the performance of Auto-PGD versus PGD with Momentum with different fixed step sizes
as a function of the iterations to show that Auto-PGD adapts automatically to a given budget of iterations and to a good step size, achieving larger (better) loss and smaller (better) robust test accuracy than PGD with momentum.
In Sec. \ref{sec:apgd_vs_pgd} we show that the same holds if we compare Auto-PGD to PGD (without momentum).
Moreover, we provide in Sec. \ref{sec:bigcomparison-apgd-pgd} a full comparison of PGD, PGD with Momentum, both with three
different step sizes, and Auto-PGD. It turns out that Auto-PGD outperforms PGD with a fixed step size showing again that the
automatic adaptation of Auto-PGD works well. 

%

\subsection{Comparison of APGD to PGD } \label{sec:apgd_vs_pgd} We repeat the experiment of Sec. \ref{sec:apgd_vs_pgdmomentum}, this time comparing the performance of \apgd{} to that of PGD for various fixed step sizes as a function
of the iterations.

The results of the comparison of \apgd{} vs PGD (without momentum) are shown in Figure \ref{fig:pgd_vs_apgd}. PGD without momentum achieves significantly smaller
loss values on MNIST and also worse robust accuracy than PGD with Momentum and in particular \apgd{}. Apart from this the behavior of PGD with fixed step size is very similar to the one of PGD with momentum where as we see that the loss and robust accuracy plateau quickly while \apgd{} manages to get still improvements.
In general, \apgd{} outperforms PGD in terms of both loss and robust accuracy.

\begin{figure*}[t] \flushleft
	{\small
	\begin{tabular}{*{2}{C{\columnwidth}}}
		\multicolumn{1}{c}{\hspace{5mm}\cite{MadEtAl2018}} & \multicolumn{1}{c}{\cite{ZhaEtAl2019}} \\
		{\scriptsize \hspace{16mm} MNIST  - $\epsilon=0.3$  \hspace{18mm} CIFAR-10  - $\epsilon=8/255$}& {\scriptsize \hspace{5mm}MNIST  - $\epsilon=0.3$\hspace{18mm} CIFAR-10  - $\epsilon=0.031$} \\
		\multicolumn{2}{l}{\rotatebox[origin=c]{90}{loss} \hfill
			\includegraphics[align=c, width=0.48\columnwidth, clip, trim=8mm 0mm 7mm 2mm]{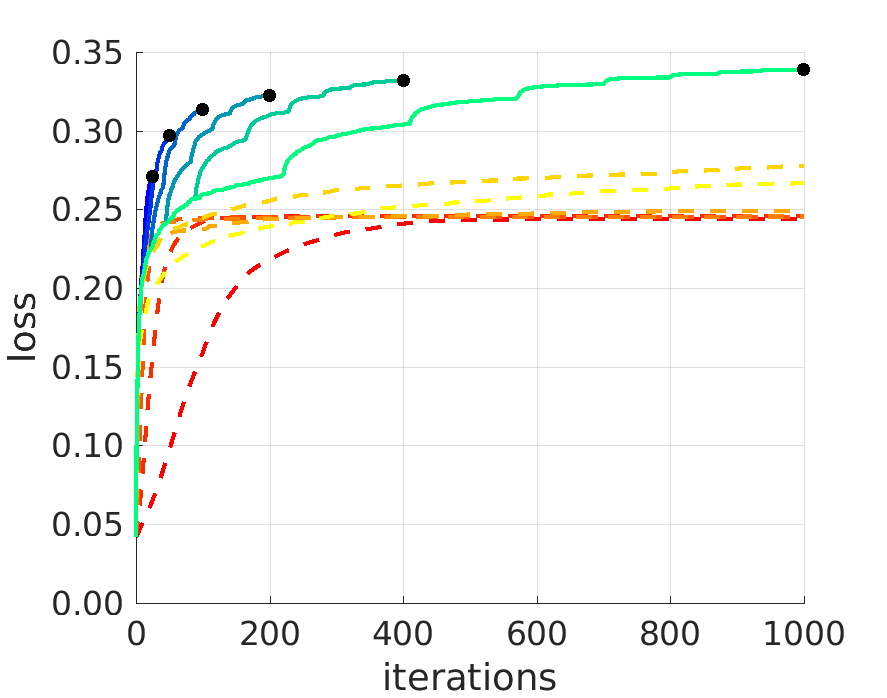} 
			\includegraphics[align=c, width=0.48\columnwidth, clip, trim=8mm 0mm 7mm 2mm]{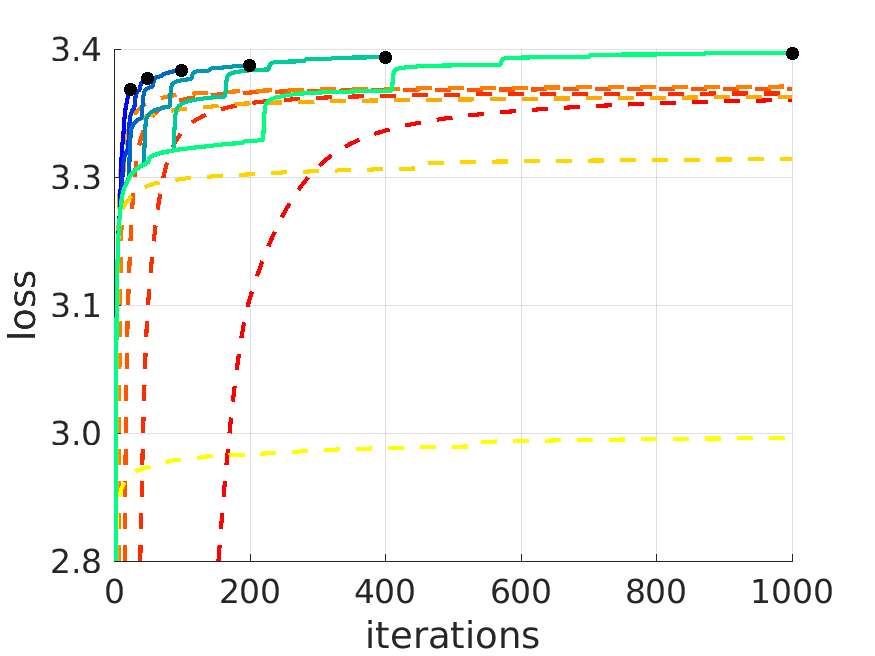}  \includegraphics[align=c, width=0.48\columnwidth, clip, trim=8mm 0mm 7mm 2mm]{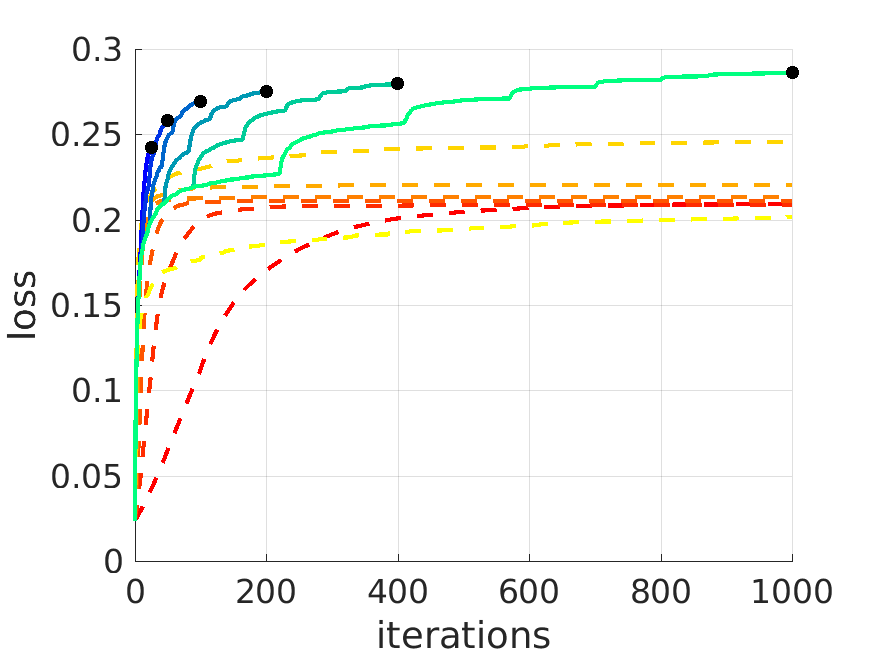} 
			\includegraphics[align=c, width=0.48\columnwidth, clip, trim=8mm 0mm 7mm 2mm]{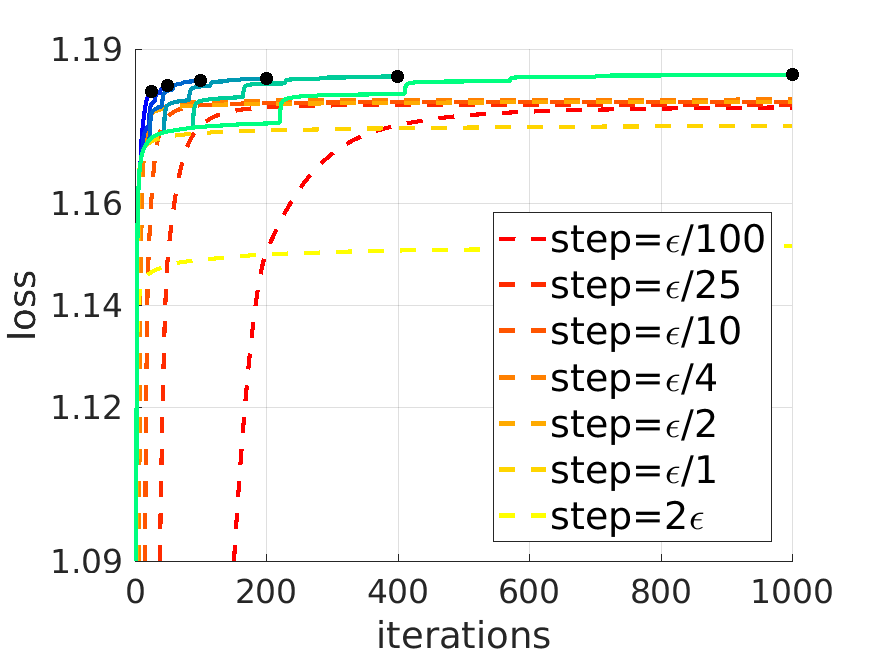}}\\
		
		\multicolumn{2}{l}{\rotatebox[origin=c]{90}{robust accuracy} \hfill
			\includegraphics[align=c, width=0.48\columnwidth, clip, trim=8mm 0mm 7mm 2mm]{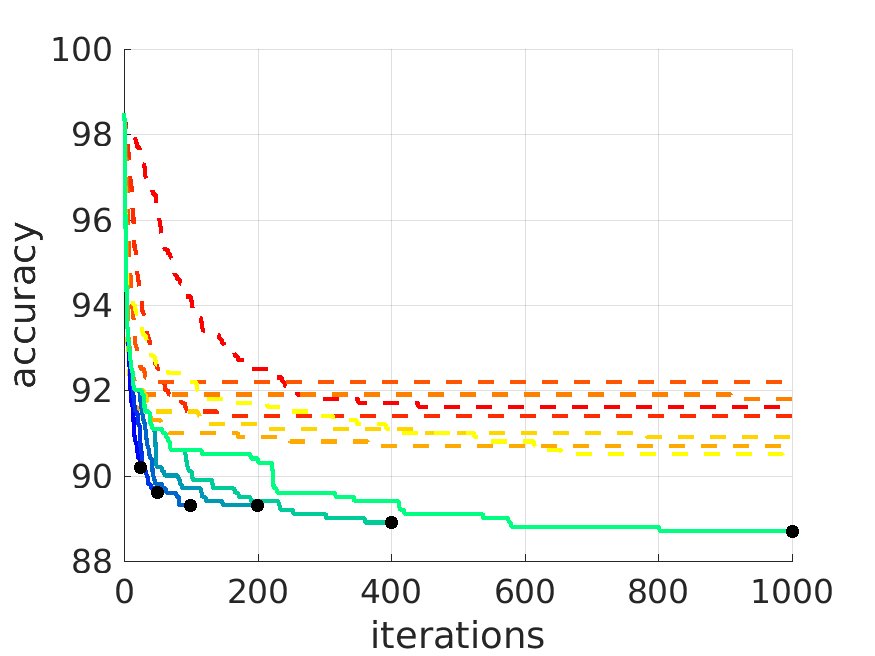} 
			\includegraphics[align=c, width=0.48\columnwidth, clip, trim=8mm 0mm 7mm 2mm]{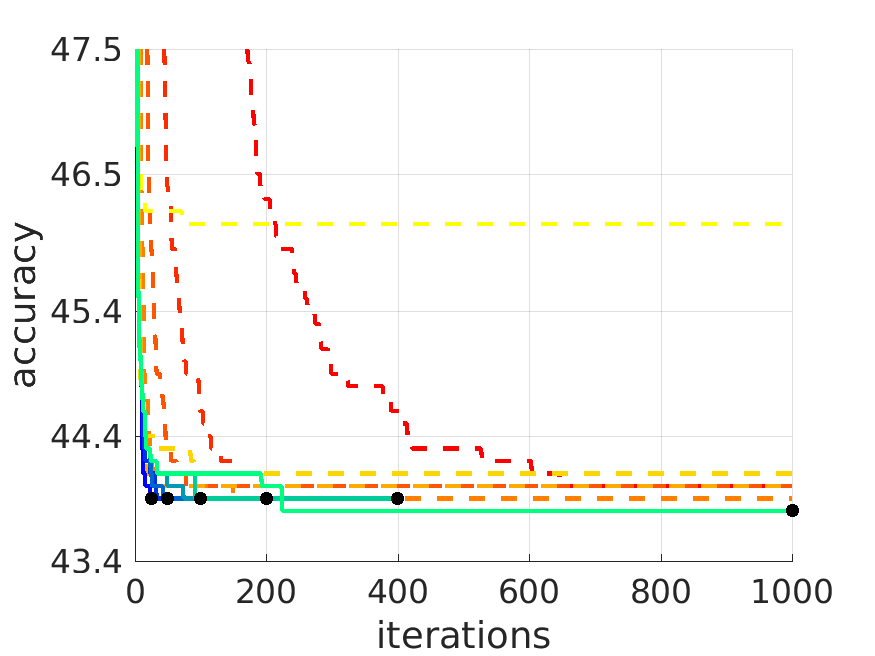}  \includegraphics[align=c, width=0.48\columnwidth, clip, trim=8mm 0mm 7mm 2mm]{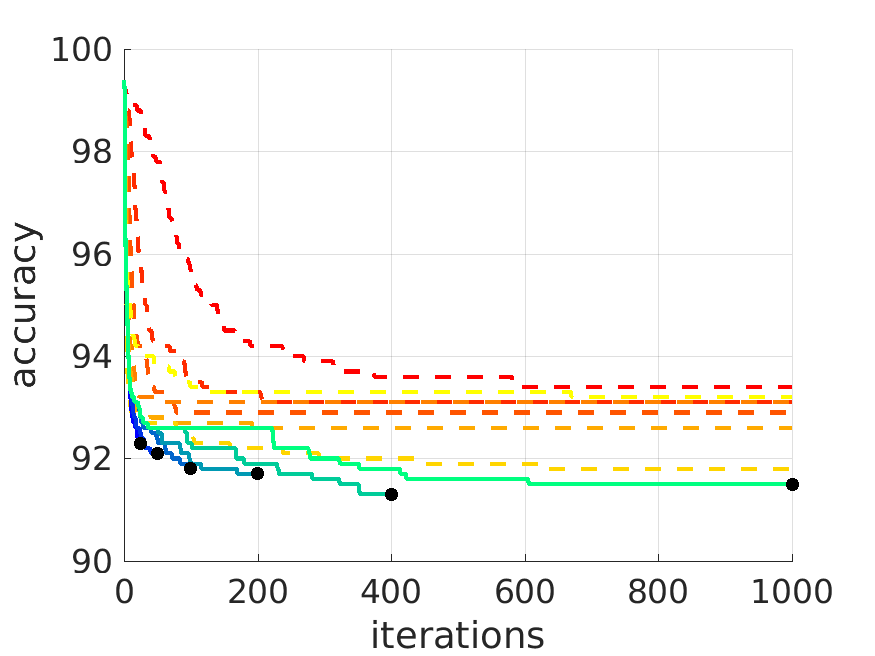}
			\includegraphics[align=c, width=0.48\columnwidth, clip, trim=8mm 0mm 7mm 2mm]{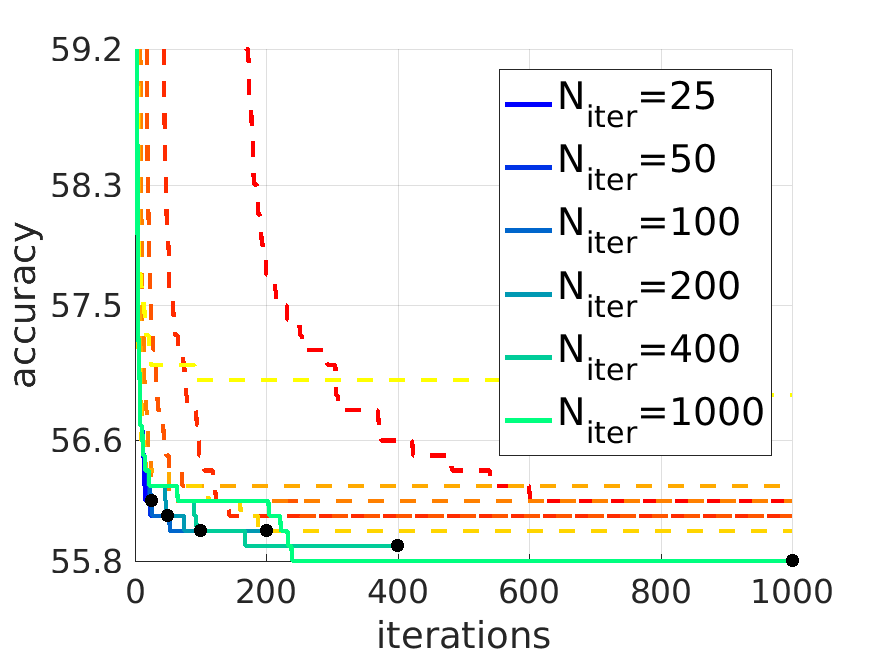}}\\
		\end{tabular}}

	\caption{\textbf{PGD vs \apgd{}:} best cross-entropy loss (top) and robust accuracy (bottom) so far found as function of iterations for the models of \cite{MadEtAl2018} and TRADES \cite{ZhaEtAl2019} for PGD (dashed lines) with different fixed step sizes (always $1000$ iterations) and \apgd{} (solid lines) with
		different budgets of iterations. \apgd{} outperforms PGD for every budget of iterations in robust accuracy.} \label{fig:pgd_vs_apgd}
\end{figure*}

\section{\Combo{}: implementation details}\label{sec:exp_details_app}
We report all the hyperparameters used in \Combo{}. For \apgd{}, we use as direction in the update step the sign of the gradient for the $l_\infty$-threat model and, as common in literature (see e.g. \cite{TsiEtAl18}), the normalized (wrt $l_2$) gradient for the $l_2$-threat model. Moreover, we set the momentum coefficient to $\alpha=0.75$, $\rho=0.75$ (Condition 1), initial step size $\iter{\eta}{0}=2\epsilon$ where $\epsilon$ is the maximum $l_p$-norm of the perturbations. For FAB we keep the standard hyperparameters according to the implementation of Advertorch\footnote{\url{https://github.com/BorealisAI/advertorch}} \cite{ding2019advertorch}. For Square Attack we follow the original code\footnote{\url{https://github.com/max-andr/square-attack}}, using as initial value for the size of the squares $p=0.8$. Moreover, we use the piecewise constant schedule for $p$ as it is suggested for a query limit of 10000 without any rescaling (although we use only up to 5000 queries).

\begin{table*}[t] \caption{\textbf{Robustness evaluation of additional adversarial defenses by \Combo{}.}
		We report clean test accuracy, the robust accuracy of the individual attacks and the combined one of \Combo{} (AA column). We also provide the robust accuracy reported 
		in the original papers and compute the difference to that of \Combo{}. If negative (in red) \Combo{} provides lower (better) robust accuracy. 
	} \label{tab:eval_additional_defenses}
	\vspace{2mm}
	\centering
		\small
		\begin{tabular}{r l R{10mm} || R{10mm} *{3}{R{10mm}} >{\columncolor[rgb]{0.9 1.0 0.9}}R{10mm}| *{2}{R{10mm}}}
			\# & paper &clean&  APGD\textsubscript{CE}& \apgd{}\SPSB{T}{DLR}& FAB\SP{T} & Square & AA & reported & reduct.\\
			\hline
			\multicolumn{10}{c}{}\\
			\multicolumn{10}{l}{\textbf{CIFAR-10} - $l_\infty$ - $\epsilon=4/255$}\\ \hline
			1 & \cite{song2018improving} & 84.81 &\textcolor{Gray}{\underline{57.43}} &\textcolor{Gray}{\underline{\textbf{56.51}}} &\textcolor{Gray}{\underline{56.86}} &\textcolor{Gray}{64.67} & 56.51 & 58.1 & \textcolor{red}{-1.59} \\ \multicolumn{10}{c}{}\\ \multicolumn{10}{l}{\textbf{CIFAR-10} - $l_\infty$ - $\epsilon=0.02$} \\ \hline 1 & \cite{pmlr-v97-pang19a} & 91.22 &\textcolor{Gray}{\underline{5.02}} &\textcolor{Gray}{\underline{\textbf{1.74}}} &\textcolor{Gray}{\underline{2.45}} &\textcolor{Gray}{\underline{29.52}} & 1.02 & 34.0 & \textcolor{red}{-32.98} \\   2 & \cite{pmlr-v97-pang19a} & 93.44 &\textcolor{Gray}{\underline{0.18}} &\textcolor{Gray}{\underline{\textbf{0.01}}} &\textcolor{Gray}{\underline{0.06}} &\textcolor{Gray}{88.55} & 0.00 & 30.4 & \textcolor{red}{-30.40}
	\end{tabular}\end{table*}

\section{Experiments}\label{sec:exps_app}
\textbf{Deterministic defenses:} We report in Table~\ref{tab:eval_additional_defenses} the results for the deterministic defenses wrt $l_\infty$ omitted in Sec.~\ref{sec:exps}, involving CIFAR-10 models with less common thresholds $\epsilon$, with the same statistics as in Table~\ref{tab:eval_def_details}. 
We add in Tables \ref{tab:deterministic_defenses_app} and \ref{tab:deterministic_defenses_app_l2} (for $l_\infty$- and $l_2$-defenses respectively) more details about the classifiers and papers considered in our evaluation: the venue where each paper appeared (we add "R" to indicate that the paper was submitted at the venue but rejected) and the source of the models, that is whether they are publicly \textit{available}, we got them via private communication from the \textit{authors} or we \textit{retrained} them according to official implementations. We also indicate the architecture of the classifiers: interestingly, larger models do not imply better robustness.

The models of \cite{pmlr-v97-pang19a} and \cite{pang2020rethinking} appears twice as they are evaluated both with and without the addition of adversarial training. The models of \cite{Wang-2019-ICCV} are the networks named "R-MOSA-LA-4" and "R-MC-LA-4".

\textbf{Randomized defenses:} Similar to the deterministic defenses we report additional information which had to be omitted in Table \ref{tab:eval_rand_def_details}. We report in Table \ref{tab:rand_def_details_app} the venue where each paper appeared and the source of the models, that is if they are publicly \textit{available} or we got them via private communication from the \textit{authors}. In Table \ref{tab:rand_def_statistics_over_runs_app} we report the mean and standard deviation of the robust accuracy achieved by each attack (note that we have a randomized defense thus the outcome is not deterministic). As mentioned in Sec. \ref{sec:exps}, to evaluate an attack we compute 5 times the classification accuracy of the target model on the same adversarial examples. For \Combo{}, we form for each test point a batch with the adversarial examples among those of our different attacks which were misclassified most often in the 5 runs, and then compute the robust accuracy of this batch.

For the models of \cite{grathwohl2020your}, only plots of robust accuracy vs size of the perturbation $\epsilon$ are available in their paper. Thus the reported values are obtained by extrapolating the values in the tables from the most recent version of the paper, that is the one available at  \url{https://arxiv.org/abs/1912.03263v2}.

\subsection{Comparison PGD vs PGD with Momentum vs \apgd{} on different losses}\label{sec:bigcomparison-apgd-pgd}
We add here more details about the comparison introduced in Sec.~\ref{sec:comparison_losses}.
In order to compare the performance of \apgd{} to that of standard PGD when testing adversarial defenses, we run PGD and PGD with Momentum on the models used in Sec. \ref{sec:exps}, with the same budget of \apgd{}, i.e. 100 iterations and 5 restarts. We use three step sizes: $\epsilon/10$, $\epsilon/4$ and $2\epsilon$, where $\epsilon$ is the threshold at which the robust accuracy is computed.
We repeat the experiment for the CE (Tables \ref{tab:app_pgd_vs_apgd_ce} and \ref{tab:app_pgd_vs_apgd_ce_l2}), CW (Tables \ref{tab:app_pgd_vs_apgd_cw_loss} and \ref{tab:app_pgd_vs_apgd_cw_loss_l2}) and DLR loss (Tables \ref{tab:app_pgd_vs_apgd_new_loss} \ref{tab:app_pgd_vs_apgd_new_loss_l2}).

We see that \apgd{} achieves with all losses the lowest (lower is better) robust accuracy in most of the cases (36 out of 49 with CE loss, 42/49 with CW loss, 39/49 with the DLR loss). Moreover, while on average PGD with Momentum and step size $\epsilon/4$ is slightly better than the other versions of PGD, the best version of PGD varies with the classifier, and in particular there is large variance among the results obtained by different step sizes. This emphasizes again why a step size free method is essential for a reliable robustness evaluation which is as much as possible independent of the user.

For the attacks on the CW and DLR loss, the only cases where PGD is more than $1\%$ better than \apgd{} consists of 
cases where the very large step size $2\epsilon$ outperforms all other step sizes. Note that the stepsize $2\epsilon$ works significantly worse 
than the smaller step sizes for most of the other models. It is likely that these defenses modify the loss landscape to make it unsuitable for standard gradient-based optimization, and an \textit{informed random search} (as PGD with such a large step size can be seen) works better. The same happens with the CE loss, where PGD yields the best results (with a non-negligible gap to \apgd{}) also on the models of \cite{kim2020sensible} and \cite{Taghanaki-2019-CVPR}. However, notice that for these classifiers optimizing the DLR loss is significantly more effective than using the CE loss. Finally, in all these cases \Combo{} achieves significantly lower robust accuracy than any version of PGD as FAB attack is less affected by this kind of gradient obfuscation.

\section{Potential failure cases of \apgd{}\textsubscript{CE}  and \apgd{}\textsubscript{DLR}}\label{sec:failure}
It is always important to understand in which cases some attacks do not work. The automatic step size selection in \apgd{} prevents a lot of the failure
cases of PGD with a fixed step size. However, clearly both losses are affected when the gradient information in general is not helpful e.g.
the defense of \cite{xiao2020enhancing} leads to a discontinuous classifier function and thus \apgd{}\textsubscript{CE}  and \apgd{}\textsubscript{DLR}
overestimate the robustness (even though they still get much lower robust accuracy than \cite{xiao2020enhancing}).
However, this particular defense shows the value of \Combo{} as in this case the black-box Square Attack which does not rely on gradient information reveals that this defense is not robust. In fact we have run
Square attack with a query limit of 10000 and 10 random restarts (instead of 5000 and 1 in the experiments) and got a robust accuracy of 8.0\% (on 500 points) compared to 20.44\% in our experiments.

\subsection{Cross-entropy loss}
As mentioned in Sec.~\ref{sec:new_loss}, for the cross-entropy loss a problem is that the loss is not invariant to rescaling of the classifier.
\comment{
This can lead to issues when one computes the gradient of the CE loss wrt $x$ which is given by 
\begin{align*} 
\nabla_x \textrm{CE}(x,y) =& -\nabla_x z_y + \sum_{i=1}^K \frac{e^{z_i}}{\sum_{j=1}^Ke^{z_j}}\nabla_x z_i\\
=&\left(-1 + p_y\right)\nabla_x z_y + \sum_{i\neq y} p_i\nabla_x z_i. 
\end{align*} 
If $p_y\approx 1$ and consequently $p_i\approx 0$ for $i\neq y$, we get $\nabla_x \textrm{CE}(x,y)\approx \mathbf{0}$, which means that maximizing the CE loss with gradient-based methods becomes ineffective \cite{PapEtAl2016a}. Notice that one can achieve $p_y\approx 1$ with a classifier $h =  \alpha g$ equivalent to $g$ (i.e. they take the same decision for every $x$) but rescaled by a constant $\alpha > 0$. Since \[\lim_{\alpha \rightarrow +\infty} \frac{e^{\alpha z_y}}{\sum_{j=1}^Ke^{\alpha z_j}} =  \begin{cases} 1 & \textrm{if}\; z_y = \maxop_i z_i \\ 0 & \textrm{else} \end{cases}, \] for sufficiently large $\alpha$ the decisions of $h$ can have arbitrary high confidence.}
The finite arithmetic
%
leads to the fact that when its gradient, in Eq.~\ref{eq:grad_ce}, is analytically $\nabla_x \textrm{CE}(x,y)\approx \mathbf{0}$, it becomes in practice
$\nabla_x \textrm{CE}(x,y)= \mathbf{0}$, since typically single precision is used and only exponents roughly in $[-127, 127]$ in the exponential function can be expressed (recently \cite{WonEtAl20} proposed even half precision).
Thus the sign of the gradient (which in principle is not affected by the magnitude of the values) becomes zero and one does not get meaningful ascent directions (known as gradient masking).

However, we highlight that maximizing the CE loss minimizes the confidence of the classifier in the correct class. This helps to test randomized defenses, when misclassification alone is not enough for a successful attack.

\subsection{Difference-of-Logits Ratio (DLR) loss}
As discussed in Sec. \ref{sec:comparison_losses} the DLR loss is the best performing loss on average when integrated into \apgd{}. 
As it is rescaling and shift-invariant it does not suffer from the issues of the cross-entropy loss. In our large scale evaluation \apgd{}\textsubscript{DLR}
is the best performing attack in the sense that it has the lowest maximal difference to best performing attack across all models we tested.
However, as discussed above \apgd{}\textsubscript{DLR} can potentially perform not very well for discontinuous classifiers as seen in the 
evaluation of \cite{xiao2020enhancing}.

%
%

%
%
%


%

%

%
%

%
%
%

%
%

%

%

\begin{table*}[p]
	\caption{\textbf{Robustness evaluation of $l_\infty$-adversarial defenses with \Combo{}.} For each model, we report
		architecture, source, venue,
		clean accuracy and combined robust accuracy given by \Combo{} (AA column). We also provide the reported robust accuracy
		in the original papers and compute the difference to the one of \Combo{}.
	} \label{tab:deterministic_defenses_app} \vspace{2mm}
	\centering
	{\centering \small 
		\setlength{\tabcolsep}{5.5pt}
		\begin{tabular}{r *{4}{l} R{10mm}|| >{\columncolor[rgb]{0.9 1.0 0.9}}R{10mm} R{10mm}| R{10mm}}
			\# & paper & model & source & venue &clean& AA& reported & reduct. \\
			\hline
			\multicolumn{9}{c}{}\\
			\multicolumn{9}{l}{\textbf{CIFAR-10} - $\epsilon=8/255$}\\
			\hline
			1 & \cite{CarEtAl19} & WideResNet-28-10 & available & NeurIPS 19 & 89.69 & 59.53 & 62.5 & \textcolor{red}{-2.97} \\   2 & \cite{AlaEtAl19} & WideResNet-106-8 & available & NeurIPS 19 & 86.46 & 56.03 & 56.30 & \textcolor{red}{-0.27} \\   3 & \cite{pmlr-v97-hendrycks19a} & WideResNet-28-10 & available & ICML 19 & 87.11 & 54.92 & 57.4 & \textcolor{red}{-2.48} \\   4 & \cite{rice2020overfitting} & WideResNet-34-20 & available & ICML 20 & 85.34 & 53.42 & 58 & \textcolor{red}{-4.58} \\   5 & \cite{qin2019adversarial} & WideResNet-40-8 & available & NeurIPS 19 & 86.28 & 52.84 & 52.81 & \textcolor{blue}{0.03} \\   6 & \cite{robustness} & ResNet-50 & available & GitHub & 87.03 & 49.25 & 53.29 & \textcolor{red}{-4.04} \\   7 & \cite{KumEtAl19} & WideResNet-34-10 & available & IJCAI 19 & 87.80 & 49.12 & 53.04 & \textcolor{red}{-3.92} \\   8 & \cite{MaoEtAl19} & WideResNet-34-10 & authors & NeurIPS 19 & 86.21 & 47.41 & 50.03 & \textcolor{red}{-2.62} \\   9 & \cite{ZhaEtAl19-yopo} & WideResNet-34-10 & retrained & NeurIPS 19 & 87.20 & 44.83 & 47.98 & \textcolor{red}{-3.15} \\   10 & \cite{MadEtAl2018} & WideResNet-34-10 & available & ICLR 18 & 87.14 & 44.04 & 47.04 & \textcolor{red}{-3.00} \\   11 & \cite{pang2020rethinking} & ResNet32 & available & ICLR 20 & 80.89 & 43.48 & 55.0 & \textcolor{red}{-11.52} \\   12 & \cite{WonEtAl20} & ResNet18 & available & ICLR 20 & 83.34 & 43.21 & 46.06 & \textcolor{red}{-2.85} \\   13 & \cite{ShaEtAl19} & WideResNet-34-10 & available & NeurIPS 19 & 86.11 & 41.47 & 46.19 & \textcolor{red}{-4.72} \\   14 & \cite{Ding2020MMA} & WideResNet-28-4 & available & ICLR 20 & 84.36 & 41.44 & 47.18 & \textcolor{red}{-5.74} \\   15 & \cite{Moosavi-Dezfooli-2019-CVPR} & WideResNet-28-10 & authors & CVPR 19 & 83.11 & 38.50 & 41.4 & \textcolor{red}{-2.90} \\   16 & \cite{ZhaWan19} & WideResNet-28-10 & available & NeurIPS 19 & 89.98 & 36.64 & 60.6 & \textcolor{red}{-23.96} \\   17 & \cite{zhang2020adversarial} & WideResNet-28-10 & available & ICLR 20 R & 90.25 & 36.45 & 68.7 & \textcolor{red}{-32.25} \\   18 & \cite{Jang-2019-ICCV} & ResNet20 & available & ICCV 19 & 78.91 & 34.95 & 37.40 & \textcolor{red}{-2.45} \\   19 & \cite{kim2020sensible} & WideResNet-34-10 & available & ICLR 20 R & 91.51 & 34.22 & 57.23 & \textcolor{red}{-23.01} \\   20 & \cite{Moosavi-Dezfooli-2019-CVPR} & ResNet18 & authors & CVPR 19 & 80.41 & 33.70 & 36.3 & \textcolor{red}{-2.60} \\   21 & \cite{Wang-2019-ICCV} & WideResNet-28-10 & available & ICCV 19 & 92.80 & 29.35 & 58.6 & \textcolor{red}{-29.25} \\   22 & \cite{Wang-2019-ICCV} & WideResNet-28-10 & available & ICCV 19 & 92.82 & 26.93 & 66.9 & \textcolor{red}{-39.97} \\   23 & \cite{Mustafa-2019-ICCV} & ResNet110 & available & ICCV 19 & 89.16 & 0.28 & 32.32 & \textcolor{red}{-32.04} \\   24 & \cite{chan2020jacobian} & WideResNet-34-10 & retrained & ICLR 20 & 93.79 & 0.26 & 15.5 & \textcolor{red}{-15.24} \\   25 & \cite{pang2020rethinking} & ResNet110 & available & ICLR 20 & 93.52 & 0.00 & 31.4 & \textcolor{red}{-31.40} \\ \multicolumn{9}{c}{}\\ \multicolumn{9}{l}{\textbf{CIFAR-10} - $l_\infty$ - $\epsilon=0.031$} \\ \hline 1 & \cite{ZhaEtAl2019} & WideResNet-34-10 & available & ICML 19 & 84.92 & 53.08 & 56.43 & \textcolor{red}{-3.35} \\   2 & \cite{AtzEtAl19} & ResNet18 & available & NeurIPS 19 & 81.30 & 40.22 & 43.17 & \textcolor{red}{-2.95} \\   3 & \cite{xiao2020enhancing} & DenseNet121 & available & ICLR 20 & 79.28 & 18.50 & 52.4 & \textcolor{red}{-33.90} \\ \multicolumn{9}{c}{}\\ \multicolumn{9}{l}{\textbf{CIFAR-10} - $l_\infty$ - $\epsilon=4/255$} \\ \hline 1 & \cite{song2018improving} & ConvNet & available & ICLR 19 & 84.81 & 56.51 & 58.1 & \textcolor{red}{-1.59} \\ \multicolumn{9}{c}{}\\ \multicolumn{9}{l}{\textbf{CIFAR-10} - $l_\infty$ - $\epsilon=0.02$} \\ \hline 1 & \cite{pmlr-v97-pang19a} & ResNet20 (x3) & available & ICML 19 & 91.22 & 1.02 & 34.0 & \textcolor{red}{-32.98} \\   2 & \cite{pmlr-v97-pang19a} & ResNet20 (x3) & available & ICML 19 & 93.44 & 0.00 & 30.4 & \textcolor{red}{-30.40} \\ \multicolumn{9}{c}{}\\ \multicolumn{9}{l}{\textbf{CIFAR-100} - $l_\infty$ - $\epsilon=8/255$} \\ \hline 1 & \cite{pmlr-v97-hendrycks19a} & WideResNet-28-10 & available & ICML 19 & 59.23 & 28.42 & 33.5 & \textcolor{red}{-5.08} \\   2 & \cite{rice2020overfitting} & PreActResNet-18 & available & ICML 20 & 53.83 & 18.95 & 28.1 & \textcolor{red}{-9.15} \\ \multicolumn{9}{c}{}\\ \multicolumn{9}{l}{\textbf{MNIST} - $l_\infty$ - $\epsilon=0.3$} \\ \hline 1 & \cite{zhang2020towards} &  & available & ICLR 20 & 98.38 & 93.96 & 96.38 & \textcolor{red}{-2.42} \\   2 & \cite{GowEtAl18} &  & available & ICCV 19 & 98.34 & 92.83 & 93.88 & \textcolor{red}{-1.05} \\   3 & \cite{ZhaEtAl2019} &  & available & ICML 19 & 99.48 & 92.81 & 95.60 & \textcolor{red}{-2.79} \\   4 & \cite{Ding2020MMA} &  & available & ICLR 20 & 98.95 & 91.40 & 92.59 & \textcolor{red}{-1.19} \\   5 & \cite{AtzEtAl19} &  & available & NeurIPS 19 & 99.35 & 90.85 & 97.35 & \textcolor{red}{-6.50} \\   6 & \cite{MadEtAl2018} &  & available & ICLR 18 & 98.53 & 88.50 & 89.62 & \textcolor{red}{-1.12} \\   7 & \cite{Jang-2019-ICCV} &  & available & ICCV 19 & 98.47 & 87.99 & 94.61 & \textcolor{red}{-6.62} \\   8 & \cite{WonEtAl20} &  & available & ICLR 20 & 98.50 & 82.93 & 88.77 & \textcolor{red}{-5.84} \\   9 & \cite{Taghanaki-2019-CVPR} &  & retrained & CVPR 19 & 98.86 & 0.00 & 64.25 & \textcolor{red}{-64.25} \\ \multicolumn{9}{c}{}\\ \multicolumn{9}{l}{\textbf{ImageNet} - $l_\infty$ - $\epsilon=4/255$} \\ \hline 1 & \cite{robustness} & ResNet-50 & available & GitHub & 63.4 & 27.6 & 33.38 & \textcolor{red}{-5.78}
	\end{tabular}}

\end{table*}

\begin{table*}[p]
	\caption{\textbf{Robustness evaluation of $l_2$-adversarial defenses with \Combo{}.} For each model, we report
		architecture, source, venue,
		clean accuracy and combined robust accuracy given by \Combo{} (AA column). We also provide the reported robust accuracy
		in the original papers and compute the difference to the one of \Combo{}.
	} \label{tab:deterministic_defenses_app_l2} \vspace{2mm}
	\centering
	{\centering \small 
		\begin{tabular}{r *{4}{l} R{10mm}|| >{\columncolor[rgb]{0.9 1.0 0.9}}R{10mm} R{10mm}| R{10mm}}
			\# & paper & model & source & venue &clean& AA& reported & reduct. \\
			\hline
			\multicolumn{9}{c}{}\\
			\multicolumn{9}{l}{\textbf{CIFAR-10} - $l_2$ - $\epsilon=0.5$}\\
			\hline
			1 & \cite{augustin2020adversarial} & ResNet-50 & authors & ECCV 20 & 91.08 & 72.91 & 73.27 & \textcolor{red}{-0.36} \\   2 & \cite{robustness} & ResNet-50 & available & GitHub & 90.83 & 69.24 & 70.11 & \textcolor{red}{-0.87} \\   3 & \cite{rice2020overfitting} & PreActResNet-18 & available & ICML 20 & 88.67 & 67.68 & 71.6 & \textcolor{red}{-3.92} \\   4 & \cite{rony2019decoupling} & WideResNet-28-10 & available & CVPR 19 & 89.05 & 66.44 & 67.6 & \textcolor{red}{-1.16} \\   5 & \cite{Ding2020MMA} & WideResNet-28-4 & available & ICLR 20 & 88.02 & 66.09 & 66.18 & \textcolor{red}{-0.09} \\ \multicolumn{9}{c}{}\\ \multicolumn{9}{l}{\textbf{ImageNet} - $l_2$ - $\epsilon=3$} \\ \hline 1 & \cite{robustness} & ResNet-50 & available & GitHub & 55.3 & 28.3 & 35.09 & \textcolor{red}{-6.79}
		\end{tabular}}\end{table*}

\begin{table*}[p]
	\caption{
	\textbf{Robustness evaluation of randomized $l_\infty$-adversarial defenses by \Combo.} For each model, we report
	architecture, source, venue,
	clean accuracy and combined robust accuracy given by \Combo{} (AA column). We also provide the reported robust accuracy
	in the original papers and compute the difference to the one of \Combo{}.}
\label{tab:rand_def_details_app}
\vspace{2mm}
\centering \small  \begin{tabular}{r *{4}{l} R{16mm}|| *{1}{>{\columncolor[rgb]{0.9 1.0 0.9}}R{16mm}} R{8mm} |R{8mm}}
		%
		\# & paper & model & source & venue &clean& \Combo{} &report. & reduct. \\
		\hline
		\multicolumn{9}{c}{}\\
		\multicolumn{9}{l}{\textbf{CIFAR-10} - $\epsilon=8/255$} \\ \hline 1 & \cite{WanEtAl19} & En\textsubscript{5}RN & authors & NeurIPS 19 & 82.39 (0.14)  & 45.56 (0.20)  & 51.48 & \textcolor{red}{-5.9} \\  2 & \cite{pmlr-v97-yang19e} & with AT & authors & ICML 19 & 84.9 (0.6)  & 26.3 (0.85)  & 52.8 & \textcolor{red}{-26.5} \\  3 & \cite{pmlr-v97-yang19e} & pure & authors & ICML 19 & 87.2 (0.3)  & 18.2 (0.82)  & 40.8 & \textcolor{red}{-22.6} \\  4 & \cite{grathwohl2020your} & JEM-10 & available & ICLR 20 & 90.99 (0.03)  & 9.92 (0.03)  & 47.6 & \textcolor{red}{-37.7} \\  5 & \cite{grathwohl2020your} & JEM-1 & available & ICLR 20 & 92.31 (0.04)  & 8.15 (0.05)  & 41.8 & \textcolor{red}{-33.6} \\  6 & \cite{grathwohl2020your} & JEM-0 & available & ICLR 20 & 92.82 (0.05)  & 6.36 (0.06)  & 19.8 & \textcolor{red}{-13.4} \\ \multicolumn{9}{c}{}\\ \multicolumn{9}{l}{\textbf{CIFAR-10} - $\epsilon=4/255$} \\ \hline 1 & \cite{grathwohl2020your} & JEM-10 & available & ICLR 20 & 91.03 (0.05)  & 47.97 (0.05)  & 72.6 & \textcolor{red}{-24.6} \\  2 & \cite{grathwohl2020your} & JEM-1 & available & ICLR 20 & 92.34 (0.04)  & 45.49 (0.04)  & 67.1 & \textcolor{red}{-21.6} \\  3 & \cite{grathwohl2020your} & JEM-0 & available & ICLR 20 & 92.82 (0.02)  & 42.55 (0.07)  & 50.8 & \textcolor{red}{-8.2} 
	\end{tabular} \end{table*}
\begin{table*}[p]
	\caption{
		\textbf{Robustness evaluation of randomized $l_\infty$-adversarial defenses by \Combo.} We  
		report mean and standard deviation over 5 runs of evaluation of the robust accuracy achieved by the adversarial examples crafted by the different attacks, together with their combination in \Combo{}.}\label{tab:rand_def_statistics_over_runs_app}
	\vspace{2mm}
	\centering \small  \begin{tabular}{r *{2}{l} || *{4}{R{16mm}} *{1}{>{\columncolor[rgb]{0.9 1.0 0.9}}R{16mm}}}
		%
		\# & paper & model & APGD\textsubscript{CE}&APGD\textsubscript{DLR} &FAB & Square& \Combo{} \\
		\hline
		\multicolumn{8}{c}{}\\
		\multicolumn{8}{l}{\textbf{CIFAR-10} - $\epsilon=8/255$} \\ \hline 1 & \cite{WanEtAl19} & En\textsubscript{5}RN & 48.81 (0.16)  & 49.37 (0.17)  & - & 78.61 (0.19)  & 45.56 (0.20)  \\  2 & \cite{pmlr-v97-yang19e} & with AT & 30.1 (0.5)  & 31.9 (0.53)  & - & - & 26.3 (0.85)  \\  3 & \cite{pmlr-v97-yang19e} & pure & 21.5 (0.6)  & 24.3 (0.98)  & - & - & 18.2 (0.82)  \\  4 & \cite{grathwohl2020your} & JEM-10 & 11.69 (0.04)  & 15.88 (0.06)  & 63.07 (0.21)  & 79.32 (0.04)  & 9.92 (0.03)  \\  5 & \cite{grathwohl2020your} & JEM-1 & 9.15 (0.02)  & 13.85 (0.05)  & 62.71 (0.17)  & 79.25 (0.14)  & 8.15 (0.05)  \\  6 & \cite{grathwohl2020your} & JEM-0 & 7.19 (0.06)  & 12.63 (0.04)  & 66.48 (0.34)  & 73.12 (0.13)  & 6.36 (0.06)  \\ \multicolumn{8}{c}{}\\ \multicolumn{8}{l}{\textbf{CIFAR-10} - $\epsilon=4/255$} \\ \hline 1 & \cite{grathwohl2020your} & JEM-10 & 49.10 (0.09)  & 52.55 (0.06)  & 78.87 (0.16)  & 89.32 (0.06)  & 47.97 (0.05)  \\  2 & \cite{grathwohl2020your} & JEM-1 & 46.08 (0.04)  & 49.71 (0.05)  & 78.93 (0.19)  & 90.17 (0.03)  & 45.49 (0.04)  \\  3 & \cite{grathwohl2020your} & JEM-0 & 42.98 (0.10)  & 47.74 (0.08)  & 82.92 (0.10)  & 89.52 (0.13)  & 42.55 (0.07) \end{tabular} \end{table*}

\begin{table*}[p]
	\caption{\textbf{Comparison PGD vs \apgd{} on the cross-entropy loss in the $l_\infty$-threat model.} We run PGD and PGD with Momentum (the same used for \apgd{}) on the models tested in Sec. \ref{sec:exps}, with three different step sizes: $\epsilon/10$, $\epsilon/4$ and $2\epsilon$.
		Similarly to \apgd{}, we use 100 iterations and 5 restarts for PGD, and report the resulting robust accuracy on the whole test set (1000 images for ImageNet).
		For each model we boldface the best attack and underline the best version of PGD (i.e. we exclude \apgd{}).}
	\label{tab:app_pgd_vs_apgd_ce}
	\vspace{2mm}
	\centering \small \begin{tabular}{r l R{10mm}|| *{3}{R{10mm}}|*{3}{R{10mm}} ||*{1}{R{10mm}}} 
& & & \multicolumn{3}{c|}{PGD} & \multicolumn{3}{c||}{PGD with Momentum} & \\
\cline{4-9}
\# & paper & clean & $\epsilon/10$ & $\epsilon/4$ & $2\epsilon$ & $\epsilon/10$ & $\epsilon/4$ & $2\epsilon$ & \apgd\\ \hline
\multicolumn{10}{c}{}\\ \multicolumn{10}{l}{\textbf{CIFAR-10} - $\epsilon=8/255$}\\ \hline 
1 & \cite{CarEtAl19} & 89.69 & 62.01  & 61.86  & 63.15  & 61.98  & \underline{61.79}  & 61.84  & \textbf{61.47}  \\  2 & \cite{AlaEtAl19} & 86.46 & 60.78  & 60.72  & 61.87  & 60.70  & \underline{60.58}  & 60.73  & \textbf{59.86}  \\  3 & \cite{pmlr-v97-hendrycks19a} & 87.11 & 57.24  & \underline{57.13}  & 57.88  & 57.28  & 57.14  & \underline{57.13}  & \textbf{57.00}  \\  4 & \cite{rice2020overfitting} & 85.34 & 56.87  & 56.82  & 57.55  & 56.88  & \underline{56.78}  & 56.81  & \textbf{56.76}  \\  5 & \cite{qin2019adversarial} & 86.28 & 55.50  & 55.48  & 56.42  & 55.50  & \underline{55.47}  & 55.62  & \textbf{55.45}  \\  6 & \cite{robustness} & 87.03 & 52.06  & 51.90  & 53.30  & 52.08  & 51.87  & \underline{51.85}  & \textbf{51.52}  \\  7 & \cite{KumEtAl19} & 87.80 & 51.85  & 51.75  & 52.95  & 51.87  & \underline{51.65}  & 51.68  & \textbf{51.56}  \\  8 & \cite{MaoEtAl19} & 86.21 & 49.71  & 49.59  & 50.69  & 49.72  & \underline{49.56}  & 49.61  & \textbf{49.39}  \\  9 & \cite{ZhaEtAl19-yopo} & 87.20 & 46.15  & 46.08  & 47.18  & 46.13  & 46.08  & \underline{46.04}  & \textbf{45.91}  \\  10 & \cite{MadEtAl2018} & 87.14 & 44.93  & 44.83  & 46.46  & 44.94  & \underline{44.81}  & 44.99  & \textbf{44.56}  \\  11 & \cite{pang2020rethinking} & 80.89 & 56.02  & 56.04  & 56.22  & 56.02  & \underline{56.01}  & 56.04  & \textbf{55.91}  \\  12 & \cite{WonEtAl20} & 83.34 & 46.10  & 45.93  & 47.45  & 46.06  & \underline{45.90}  & 46.03  & \textbf{45.60}  \\  13 & \cite{ShaEtAl19} & 86.11 & 44.45  & 45.60  & 48.83  & \underline{43.63}  & 44.50  & 47.07  & \textbf{43.30}  \\  14 & \cite{Ding2020MMA} & 84.36 & 50.57  & 50.36  & 50.16  & 50.58  & 50.41  & \underline{49.66}  & \textbf{49.36}  \\  15 & \cite{Moosavi-Dezfooli-2019-CVPR} & 83.11 & 41.76  & 41.71  & 42.33  & 41.70  & 41.68  & \underline{\textbf{41.58}}  & 41.59  \\  16 & \cite{ZhaWan19} & 89.98 & 67.38  & 66.79  & \underline{\textbf{57.31}}  & 67.40  & 66.50  & 61.15  & 62.03  \\  17 & \cite{zhang2020adversarial} & 90.25 & 71.69  & 71.42  & \underline{\textbf{62.52}}  & 71.77  & 71.48  & 68.88  & 69.36  \\  18 & \cite{Jang-2019-ICCV} & 78.91 & 38.00  & 37.88  & 39.72  & 37.90  & \underline{37.73}  & 37.90  & \textbf{37.40}  \\  19 & \cite{kim2020sensible} & 91.51 & 59.16  & 58.24  & \underline{\textbf{51.29}}  & 59.12  & 58.23  & 54.17  & 54.80  \\  20 & \cite{Moosavi-Dezfooli-2019-CVPR} & 80.41 & 36.86  & 36.74  & 37.92  & 36.81  & 36.74  & \underline{36.61}  & \textbf{36.53}  \\  21 & \cite{Wang-2019-ICCV} & 92.80 & 61.92  & 60.61  & \underline{\textbf{54.96}}  & 62.05  & 60.66  & 56.89  & 57.19  \\  22 & \cite{Wang-2019-ICCV} & 92.82 & 69.63  & 69.10  & \underline{\textbf{61.03}}  & 69.60  & 69.06  & 67.71  & 67.77  \\  23 & \cite{Mustafa-2019-ICCV} & 89.16 & 17.04  & 13.79  & \underline{\textbf{4.39}}  & 17.06  & 13.33  & 5.57  & 4.48  \\  24 & \cite{chan2020jacobian} & 93.79 & \underline{\textbf{1.89}}  & \underline{\textbf{1.89}}  & 9.67  & 1.91  & 1.90  & 5.93  & 1.90  \\  25 & \cite{pang2020rethinking} & 93.52 & 85.72  & 85.73  & \underline{85.67}  & 85.72  & 85.72  & 86.23  & \textbf{85.58}  \\ \multicolumn{10}{c}{}\\ \multicolumn{10}{l}{\textbf{CIFAR-10} - $\epsilon=0.031$} \\ \hline 1 & \cite{ZhaEtAl2019} & 84.92 & 55.28  & 55.24  & 56.24  & 55.25  & \underline{55.21}  & 55.30  & \textbf{55.08}  \\  2 & \cite{AtzEtAl19} & 81.30 & \underline{\textbf{78.94}}  & \underline{\textbf{78.94}}  & \underline{\textbf{78.94}}  & \underline{\textbf{78.94}}  & \underline{\textbf{78.94}}  & \underline{\textbf{78.94}}  & \textbf{78.94}  \\  3 & \cite{xiao2020enhancing} & 79.28 & 36.32  & 36.03  & \underline{34.19}  & 36.27  & 36.04  & 35.97  & \textbf{32.38}  \\ \multicolumn{10}{c}{}\\ \multicolumn{10}{l}{\textbf{CIFAR-10} - $\epsilon=4/255$} \\ \hline 1 & \cite{song2018improving} & 84.81 & 57.42  & \underline{\textbf{57.40}}  & 57.60  & 57.42  & 57.41  & 57.43  & 57.43  \\ \multicolumn{10}{c}{}\\ \multicolumn{10}{l}{\textbf{CIFAR-10} - $\epsilon=0.02$} \\ \hline 1 & \cite{pmlr-v97-pang19a} & 91.22 & 7.27  & 5.96  & 17.23  & 7.00  & \underline{5.44}  & 9.52  & \textbf{3.53}  \\  2 & \cite{pmlr-v97-pang19a} & 93.44 & 0.36  & 0.24  & 3.37  & 0.41  & \underline{0.17}  & 1.22  & \textbf{0.04}  \\ \multicolumn{10}{c}{}\\ \multicolumn{10}{l}{\textbf{CIFAR-100} - $\epsilon=8/255$} \\ \hline 1 & \cite{pmlr-v97-hendrycks19a} & 59.23 & 33.01  & 32.91  & 33.17  & 33.02  & 32.83  & \underline{\textbf{32.76}}  & 32.83  \\  2 & \cite{rice2020overfitting} & 53.83 & 20.61  & 20.57  & 21.08  & 20.57  & 20.52  & \underline{20.47}  & \textbf{20.32}  \\ \multicolumn{10}{c}{}\\ \multicolumn{10}{l}{\textbf{MNIST} - $\epsilon=0.3$} \\ \hline 1 & \cite{zhang2020towards} & 98.38 & 95.25  & 95.23  & 95.21  & 95.06  & \underline{94.98}  & 94.99  & \textbf{94.58}  \\  2 & \cite{GowEtAl18} & 98.34 & 94.67  & 94.58  & 94.39  & 94.38  & 94.23  & \underline{94.19}  & \textbf{93.81}  \\  3 & \cite{ZhaEtAl2019} & 99.48 & 94.07  & 94.10  & 95.27  & 93.97  & \underline{93.80}  & 94.32  & \textbf{93.14}  \\  4 & \cite{Ding2020MMA} & 98.95 & 94.44  & 94.42  & 95.96  & 94.04  & \underline{93.53}  & 95.43  & \textbf{93.51}  \\  5 & \cite{AtzEtAl19} & 99.35 & \underline{\textbf{98.79}}  & 98.83  & 98.94  & \underline{\textbf{98.79}}  & 98.83  & 98.93  & \textbf{98.79}  \\  6 & \cite{MadEtAl2018} & 98.53 & 90.75  & 90.88  & 91.97  & 90.37  & \underline{90.29}  & 91.03  & \textbf{89.40}  \\  7 & \cite{Jang-2019-ICCV} & 98.47 & 93.05  & 93.66  & 95.34  & 92.73  & \underline{92.47}  & 94.47  & \textbf{92.45}  \\  8 & \cite{WonEtAl20} & 98.50 & 87.16  & 87.40  & 90.19  & 86.51  & \underline{86.30}  & 87.23  & \textbf{84.74}  \\  9 & \cite{Taghanaki-2019-CVPR} & 98.86 & 20.61  & 20.27  & \underline{\textbf{20.23}}  & 20.56  & 20.27  & 20.43  & 23.83  \\ \multicolumn{10}{c}{}\\ \multicolumn{10}{l}{\textbf{ImageNet} - $\epsilon=4/255$} \\ \hline 1 & \cite{robustness} & 63.4 & 31.5  & \underline{31.4}  & 34.0  & 31.5  & \underline{31.4}  & 31.9  & \textbf{30.9}
\end{tabular}\end{table*}

\begin{table*}[p]
	\caption{\textbf{Comparison PGD vs \apgd{} on the CW loss in the $l_\infty$-threat model.} We run PGD and PGD with Momentum (the same used for \apgd{}) on the models tested in Sec. \ref{sec:exps}, with three different step sizes: $\epsilon/10$, $\epsilon/4$ and $2\epsilon$.
		Similarly to \apgd{}, we use 100 iterations and 5 restarts for PGD, and report the resulting robust accuracy on the whole test set (1000 images for ImageNet).
		For each model we boldface the best attack and underline the best version of PGD (i.e. we exclude \apgd{}).}\vspace{2mm}\label{tab:app_pgd_vs_apgd_cw_loss}
	\centering \small \begin{tabular}{r l R{10mm}|| *{3}{R{10mm}}|*{3}{R{10mm}} || *{1}{R{10mm}}} 
		& & & \multicolumn{3}{c|}{PGD} & \multicolumn{3}{c||}{PGD with Momentum} & \\
		\cline{4-9}
		\# & paper & clean & $\epsilon/10$ & $\epsilon/4$ & $2\epsilon$ & $\epsilon/10$ & $\epsilon/4$ & $2\epsilon$ & \apgd\\ \hline
		\multicolumn{10}{c}{}\\ \multicolumn{10}{l}{\textbf{CIFAR-10} - $\epsilon=8/255$}\\
		\hline		
		1 & \cite{CarEtAl19} & 89.69 & 60.73  & 60.72  & 62.20  & 60.73  & \underline{60.69}  & 61.02  & \textbf{60.48}  \\  2 & \cite{AlaEtAl19} & 86.46 & 61.78  & 61.85  & 63.23  & 61.69  & \underline{61.61}  & 62.10  & \textbf{61.13}  \\  3 & \cite{pmlr-v97-hendrycks19a} & 87.11 & 56.44  & 56.42  & 57.15  & 56.45  & 56.41  & \underline{56.39}  & \textbf{56.20}  \\  4 & \cite{rice2020overfitting} & 85.34 & 55.24  & 55.21  & 56.04  & 55.23  & \underline{55.19}  & 55.26  & \textbf{55.07}  \\  5 & \cite{qin2019adversarial} & 86.28 & \underline{55.10}  & \underline{55.10}  & 55.95  & \underline{55.10}  & 55.11  & 55.22  & \textbf{54.99}  \\  6 & \cite{robustness} & 87.03 & 52.24  & 52.17  & 53.61  & 52.26  & \underline{52.16}  & 52.36  & \textbf{51.94}  \\  7 & \cite{KumEtAl19} & 87.80 & 51.13  & 51.07  & 52.67  & 51.14  & \underline{51.03}  & 51.22  & \textbf{50.88}  \\  8 & \cite{MaoEtAl19} & 86.21 & 49.88  & \underline{49.84}  & 51.18  & 49.88  & \underline{49.84}  & 50.02  & \textbf{49.71}  \\  9 & \cite{ZhaEtAl19-yopo} & 87.20 & 47.13  & 47.09  & 48.27  & 47.11  & \underline{47.06}  & 47.08  & \textbf{46.89}  \\  10 & \cite{MadEtAl2018} & 87.14 & 45.98  & 45.82  & 47.47  & 45.95  & \underline{45.77}  & 46.11  & \textbf{45.67}  \\  11 & \cite{pang2020rethinking} & 80.89 & 44.40  & 44.41  & 45.83  & 44.41  & \underline{44.36}  & 44.66  & \textbf{44.11}  \\  12 & \cite{WonEtAl20} & 83.34 & 46.04  & 45.94  & 47.72  & 46.03  & \underline{45.93}  & 46.26  & \textbf{45.66}  \\  13 & \cite{ShaEtAl19} & 86.11 & 44.89  & 45.93  & 49.27  & \underline{44.16}  & 45.10  & 47.43  & \textbf{43.90}  \\  14 & \cite{Ding2020MMA} & 84.36 & 51.60  & 51.44  & 51.09  & 51.59  & 51.29  & \underline{50.63}  & \textbf{50.25}  \\  15 & \cite{Moosavi-Dezfooli-2019-CVPR} & 83.11 & 40.21  & 40.18  & 40.93  & 40.20  & \underline{40.16}  & 40.22  & \textbf{40.10}  \\  16 & \cite{ZhaWan19} & 89.98 & 55.98  & 55.30  & \underline{\textbf{48.59}}  & 55.99  & 55.27  & 50.91  & 51.65  \\  17 & \cite{zhang2020adversarial} & 90.25 & 67.00  & 66.61  & \underline{\textbf{57.87}}  & 67.02  & 66.67  & 64.19  & 64.46  \\  18 & \cite{Jang-2019-ICCV} & 78.91 & 36.85  & \underline{36.81}  & 39.32  & 36.85  & 36.83  & 37.30  & \textbf{36.52}  \\  19 & \cite{kim2020sensible} & 91.51 & 56.85  & 55.90  & \underline{\textbf{50.26}}  & 56.94  & 55.92  & 52.54  & 53.04  \\  20 & \cite{Moosavi-Dezfooli-2019-CVPR} & 80.41 & 35.38  & 35.37  & 36.86  & 35.38  & \underline{35.35}  & 35.48  & \textbf{35.29}  \\  21 & \cite{Wang-2019-ICCV} & 92.80 & 59.93  & 58.42  & \underline{\textbf{52.56}}  & 60.01  & 58.46  & 54.51  & 54.77  \\  22 & \cite{Wang-2019-ICCV} & 92.82 & 66.49  & 65.86  & \underline{\textbf{56.71}}  & 66.60  & 65.76  & 63.53  & 63.80  \\  23 & \cite{Mustafa-2019-ICCV} & 89.16 & 18.27  & 14.75  & \underline{\textbf{4.42}}  & 18.44  & 14.66  & 5.46  & 4.73  \\  24 & \cite{chan2020jacobian} & 93.79 & 1.32  & \underline{1.28}  & 5.51  & 1.30  & 1.29  & 1.29  & \textbf{1.23}  \\  25 & \cite{pang2020rethinking} & 93.52 & 1.88  & 0.54  & 1.42  & 1.59  & 0.53  & \underline{0.47}  & \textbf{0.11}  \\ \multicolumn{10}{c}{}\\ \multicolumn{10}{l}{\textbf{CIFAR-10} - $\epsilon=0.031$} \\ \hline 1 & \cite{ZhaEtAl2019} & 84.92 & 54.05  & 54.06  & 55.16  & 54.08  & \underline{54.03}  & 54.22  & \textbf{53.90}  \\  2 & \cite{AtzEtAl19} & 81.30 & 40.37  & 40.28  & 42.30  & 40.36  & \underline{40.21}  & 40.56  & \textbf{40.05}  \\  3 & \cite{xiao2020enhancing} & 79.28 & 35.77  & 35.53  & \underline{34.04}  & 36.14  & 35.31  & 35.76  & \textbf{32.09}  \\ \multicolumn{10}{c}{}\\ \multicolumn{10}{l}{\textbf{CIFAR-10} - $\epsilon=4/255$} \\ \hline 1 & \cite{song2018improving} & 84.81 & 57.54  & 57.53  & 57.78  & 57.52  & 57.53  & \underline{57.51}  & \textbf{57.49}  \\ \multicolumn{10}{c}{}\\ \multicolumn{10}{l}{\textbf{CIFAR-10} - $\epsilon=0.02$} \\ \hline 1 & \cite{pmlr-v97-pang19a} & 91.22 & 7.12  & 6.09  & 16.15  & 6.87  & \underline{5.57}  & 9.49  & \textbf{3.82}  \\  2 & \cite{pmlr-v97-pang19a} & 93.44 & 0.47  & 0.26  & 3.13  & 0.42  & \underline{0.13}  & 1.22  & \textbf{0.12}  \\ \multicolumn{10}{c}{}\\ \multicolumn{10}{l}{\textbf{CIFAR-100} - $\epsilon=8/255$} \\ \hline 1 & \cite{pmlr-v97-hendrycks19a} & 59.23 & 30.64  & 30.61  & 31.19  & 30.63  & 30.61  & \underline{30.56}  & \textbf{30.45}  \\  2 & \cite{rice2020overfitting} & 53.83 & 20.22  & \underline{20.21}  & 21.06  & 20.22  & 20.22  & 20.35  & \textbf{19.98}  \\ \multicolumn{10}{c}{}\\ \multicolumn{10}{l}{\textbf{MNIST} - $\epsilon=0.3$} \\ \hline 1 & \cite{zhang2020towards} & 98.38 & 95.22  & 95.21  & 95.17  & 95.10  & \underline{94.97}  & 95.04  & \textbf{94.61}  \\  2 & \cite{GowEtAl18} & 98.34 & 94.63  & 94.48  & 94.39  & 94.49  & \underline{94.12}  & 94.23  & \textbf{93.58}  \\  3 & \cite{ZhaEtAl2019} & 99.48 & 94.40  & 94.35  & 95.23  & 94.18  & \underline{94.04}  & 94.33  & \textbf{93.35}  \\  4 & \cite{Ding2020MMA} & 98.95 & 94.49  & 94.52  & 95.85  & 93.98  & \underline{93.81}  & 95.38  & \textbf{93.53}  \\  5 & \cite{AtzEtAl19} & 99.35 & 94.43  & 94.69  & 95.98  & 94.03  & \underline{93.74}  & 95.21  & \textbf{92.98}  \\  6 & \cite{MadEtAl2018} & 98.53 & 90.82  & 90.96  & 91.94  & 90.39  & \underline{90.30}  & 91.07  & \textbf{89.39}  \\  7 & \cite{Jang-2019-ICCV} & 98.47 & 93.28  & 93.64  & 95.26  & \underline{92.68}  & 92.71  & 94.49  & \textbf{92.40}  \\  8 & \cite{WonEtAl20} & 98.50 & 87.32  & 87.55  & 90.20  & 86.74  & \underline{86.62}  & 87.25  & \textbf{84.97}  \\  9 & \cite{Taghanaki-2019-CVPR} & 98.86 & \underline{\textbf{0.00}}  & \underline{\textbf{0.00}}  & \underline{\textbf{0.00}}  & \underline{\textbf{0.00}}  & \underline{\textbf{0.00}}  & \underline{\textbf{0.00}}  & \textbf{0.00}  \\ \multicolumn{10}{c}{}\\ \multicolumn{10}{l}{\textbf{ImageNet} - $\epsilon=4/255$} \\ \hline 1 & \cite{robustness} & 63.4 & 31.7  & \underline{31.6}  & 34.5  & 31.7  & 31.7  & 32.5  & \textbf{31.5}
	\end{tabular}\end{table*}

\begin{table*}[p]
	\caption{\textbf{Comparison PGD vs \apgd{} on the DLR loss in the $l_\infty$-threat model.} We run PGD and PGD with Momentum (the same used for \apgd{}) on the models tested in Sec. \ref{sec:exps}, with three different step sizes: $\epsilon/10$, $\epsilon/4$ and $2\epsilon$.
		Similarly to \apgd{}, we use 100 iterations and 5 restarts for PGD, and report the resulting robust accuracy on the whole test set (1000 images for ImageNet).
		For each model we boldface the best attack and underline the best version of PGD (i.e. we exclude \apgd{}).}\vspace{2mm}\label{tab:app_pgd_vs_apgd_new_loss}
	\centering \small \begin{tabular}{r l R{10mm}|| *{3}{R{10mm}}|*{3}{R{10mm}} || *{1}{R{10mm}}} 
		& & & \multicolumn{3}{c|}{PGD} & \multicolumn{3}{c||}{PGD with Momentum} & \\
		\cline{4-9}
		\# & paper & clean & $\epsilon/10$ & $\epsilon/4$ & $2\epsilon$ & $\epsilon/10$ & $\epsilon/4$ & $2\epsilon$ & \apgd\\ \hline
		\multicolumn{10}{c}{}\\ \multicolumn{10}{l}{\textbf{CIFAR-10} - $\epsilon=8/255$}\\
		\hline
		1 & \cite{CarEtAl19} & 89.69 & 60.98  & 60.88  & 62.41  & 60.93  & \underline{60.82}  & 61.25  & \textbf{60.64}  \\  2 & \cite{AlaEtAl19} & 86.46 & 63.27  & 63.15  & 64.71  & 63.16  & \underline{63.00}  & 63.55  & \textbf{62.03}  \\  3 & \cite{pmlr-v97-hendrycks19a} & 87.11 & 57.27  & \underline{57.18}  & 57.98  & 57.27  & \underline{57.18}  & 57.22  & \textbf{56.96}  \\  4 & \cite{rice2020overfitting} & 85.34 & 55.89  & 55.86  & 56.75  & 55.90  & \underline{55.85}  & 55.93  & \textbf{55.72}  \\  5 & \cite{qin2019adversarial} & 86.28 & 55.73  & 55.70  & 56.47  & 55.73  & \underline{55.68}  & 55.70  & \textbf{55.46}  \\  6 & \cite{robustness} & 87.03 & 53.13  & 53.04  & 54.41  & 53.09  & \underline{52.95}  & 53.29  & \textbf{52.62}  \\  7 & \cite{KumEtAl19} & 87.80 & 51.97  & 51.92  & 53.58  & 51.98  & \underline{51.89}  & 52.11  & \textbf{51.68}  \\  8 & \cite{MaoEtAl19} & 86.21 & 50.79  & 50.57  & 52.03  & 50.74  & \underline{50.52}  & 50.79  & \textbf{50.33}  \\  9 & \cite{ZhaEtAl19-yopo} & 87.20 & 47.70  & 47.58  & 48.84  & 47.71  & 47.57  & \underline{47.53}  & \textbf{47.33}  \\  10 & \cite{MadEtAl2018} & 87.14 & 46.49  & \underline{46.30}  & 48.13  & 46.51  & \underline{46.30}  & 46.58  & \textbf{46.03}  \\  11 & \cite{pang2020rethinking} & 80.89 & 44.97  & 44.91  & 46.36  & 44.96  & \underline{44.87}  & 45.21  & \textbf{44.56}  \\  12 & \cite{WonEtAl20} & 83.34 & 47.20  & 47.05  & 48.62  & 47.18  & \underline{47.01}  & 47.22  & \textbf{46.64}  \\  13 & \cite{ShaEtAl19} & 86.11 & 45.46  & 46.43  & 50.03  & \underline{44.74}  & 45.51  & 47.89  & \textbf{44.56}  \\  14 & \cite{Ding2020MMA} & 84.36 & 51.67  & 51.44  & 51.46  & 51.68  & 51.39  & \underline{50.93}  & \textbf{50.32}  \\  15 & \cite{Moosavi-Dezfooli-2019-CVPR} & 83.11 & 40.36  & 40.33  & 41.29  & 40.37  & \underline{40.32}  & 40.43  & \textbf{40.29}  \\  16 & \cite{ZhaWan19} & 89.98 & 55.69  & 54.73  & \underline{\textbf{46.57}}  & 55.68  & 54.68  & 48.53  & 48.96  \\  17 & \cite{zhang2020adversarial} & 90.25 & 59.16  & 56.77  & \underline{\textbf{46.72}}  & 59.13  & 56.61  & 49.07  & 49.43  \\  18 & \cite{Jang-2019-ICCV} & 78.91 & 37.41  & 37.30  & 39.77  & 37.38  & \underline{37.27}  & 37.69  & \textbf{37.01}  \\  19 & \cite{kim2020sensible} & 91.51 & 52.42  & 51.41  & \underline{\textbf{48.09}}  & 52.43  & 51.30  & 48.51  & 48.41  \\  20 & \cite{Moosavi-Dezfooli-2019-CVPR} & 80.41 & 35.57  & 35.53  & 37.25  & 35.57  & \underline{35.52}  & 35.68  & \textbf{35.47}  \\  21 & \cite{Wang-2019-ICCV} & 92.80 & 61.40  & 53.79  & \underline{\textbf{38.23}}  & 61.31  & 53.27  & 39.16  & 40.69  \\  22 & \cite{Wang-2019-ICCV} & 92.82 & 54.84  & 49.01  & \underline{\textbf{35.15}}  & 54.83  & 48.62  & 36.05  & 36.72  \\  23 & \cite{Mustafa-2019-ICCV} & 89.16 & 22.16  & 17.10  & \underline{4.90}  & 22.18  & 17.42  & 5.57  & \textbf{4.54}  \\  24 & \cite{chan2020jacobian} & 93.79 & 9.52  & 7.11  & 10.41  & 7.56  & 5.38  & \underline{5.30}  & \textbf{1.20}  \\  25 & \cite{pang2020rethinking} & 93.52 & 1.94  & \underline{\textbf{0.47}}  & 1.75  & 2.06  & 0.58  & 0.75  & 0.49  \\ \multicolumn{10}{c}{}\\ \multicolumn{10}{l}{\textbf{CIFAR-10} - $\epsilon=0.031$} \\ \hline 1 & \cite{ZhaEtAl2019} & 84.92 & 54.25  & 54.20  & 55.28  & 54.23  & \underline{54.16}  & 54.34  & \textbf{54.04}  \\  2 & \cite{AtzEtAl19} & 81.30 & 57.45  & 51.99  & 44.62  & 56.19  & 50.03  & \underline{\textbf{44.45}}  & 44.50  \\  3 & \cite{xiao2020enhancing} & 79.28 & 37.74  & 36.33  & \underline{33.46}  & 37.49  & 36.92  & 35.80  & \textbf{31.27}  \\ \multicolumn{10}{c}{}\\ \multicolumn{10}{l}{\textbf{CIFAR-10} - $\epsilon=4/255$} \\ \hline 1 & \cite{song2018improving} & 84.81 & 57.85  & 57.83  & 58.07  & 57.84  & 57.83  & \underline{57.81}  & \textbf{57.79}  \\ \multicolumn{10}{c}{}\\ \multicolumn{10}{l}{\textbf{CIFAR-10} - $\epsilon=0.02$} \\ \hline 1 & \cite{pmlr-v97-pang19a} & 91.22 & 15.83  & 13.33  & 18.98  & 15.23  & \underline{12.38}  & 16.31  & \textbf{8.61}  \\  2 & \cite{pmlr-v97-pang19a} & 93.44 & 1.53  & 0.98  & 3.58  & 1.25  & \underline{0.68}  & 2.59  & \textbf{0.26}  \\ \multicolumn{10}{c}{}\\ \multicolumn{10}{l}{\textbf{CIFAR-100} - $\epsilon=8/255$} \\ \hline 1 & \cite{pmlr-v97-hendrycks19a} & 59.23 & 32.05  & 31.91  & 32.14  & 32.03  & 31.92  & \underline{31.70}  & \textbf{31.68}  \\  2 & \cite{rice2020overfitting} & 53.83 & 20.41  & 20.34  & 21.41  & 20.40  & \underline{20.33}  & 20.47  & \textbf{20.20}  \\ \multicolumn{10}{c}{}\\ \multicolumn{10}{l}{\textbf{MNIST} - $\epsilon=0.3$} \\ \hline 1 & \cite{zhang2020towards} & 98.38 & 95.65  & 95.52  & 95.26  & 95.41  & 95.17  & \underline{95.14}  & \textbf{94.82}  \\  2 & \cite{GowEtAl18} & 98.34 & 94.93  & 94.73  & 94.28  & 94.72  & 94.45  & \underline{94.22}  & \textbf{93.87}  \\  3 & \cite{ZhaEtAl2019} & 99.48 & 94.84  & 94.78  & 95.53  & 94.63  & \underline{94.50}  & 94.79  & \textbf{93.89}  \\  4 & \cite{Ding2020MMA} & 98.95 & 94.69  & 94.78  & 95.80  & 94.16  & \underline{94.00}  & 95.40  & \textbf{93.86}  \\  5 & \cite{AtzEtAl19} & 99.35 & 94.70  & 94.99  & 96.32  & 94.10  & \underline{\textbf{93.95}}  & 95.53  & 94.54  \\  6 & \cite{MadEtAl2018} & 98.53 & 91.08  & 91.21  & 92.15  & 90.65  & \underline{90.51}  & 91.51  & \textbf{89.74}  \\  7 & \cite{Jang-2019-ICCV} & 98.47 & 93.19  & 93.79  & 94.74  & \underline{92.72}  & 92.99  & 94.49  & \textbf{92.15}  \\  8 & \cite{WonEtAl20} & 98.50 & 87.58  & 87.65  & 90.40  & 86.99  & \underline{86.87}  & 87.74  & \textbf{85.39}  \\  9 & \cite{Taghanaki-2019-CVPR} & 98.86 & \underline{\textbf{0.00}}  & \underline{\textbf{0.00}}  & \underline{\textbf{0.00}}  & \underline{\textbf{0.00}}  & \underline{\textbf{0.00}}  & \underline{\textbf{0.00}}  & \textbf{0.00}  \\ \multicolumn{10}{c}{}\\ \multicolumn{10}{l}{\textbf{ImageNet} - $\epsilon=4/255$} \\ \hline 1 & \cite{robustness} & 63.4 & \underline{32.1}  & \underline{32.1}  & 35.0  & \underline{32.1}  & \underline{32.1}  & 32.8  & \textbf{32.0}
\end{tabular} \end{table*}
\begin{table*}[p]
	\caption{\textbf{Comparison PGD vs \apgd{} on the cross-entropy loss in the $l_2$-threat model.} We run PGD and PGD with Momentum (the same used for \apgd{}) on the models tested in Sec. \ref{sec:exps}, with three different step sizes: $\epsilon/10$, $\epsilon/4$ and $2\epsilon$.
		Similarly to \apgd{}, we use 100 iterations and 5 restarts for PGD, and report the resulting robust accuracy on the whole test set (1000 images for ImageNet).
		For each model we boldface the best attack and underline the best version of PGD (i.e. we exclude \apgd{}).}
	\label{tab:app_pgd_vs_apgd_ce_l2}
	\vspace{2mm}
	\centering \small \begin{tabular}{r l R{10mm}|| *{3}{R{10mm}}|*{3}{R{10mm}} ||*{1}{R{10mm}}} 
		& & & \multicolumn{3}{c|}{PGD} & \multicolumn{3}{c||}{PGD with Momentum} & \\
		\cline{4-9}
		\# & paper & clean & $\epsilon/10$ & $\epsilon/4$ & $2\epsilon$ & $\epsilon/10$ & $\epsilon/4$ & $2\epsilon$ & \apgd\\ \hline
		\multicolumn{10}{c}{}\\ \multicolumn{10}{l}{\textbf{CIFAR-10} - $\epsilon=0.5$}\\ \hline 
		1 & \cite{augustin2020adversarial} & 91.08 & 74.81  & 74.74  & \underline{\textbf{74.60}}  & 74.81  & 74.75  & 74.63  & 74.65  \\  2 & \cite{robustness} & 90.83 & 69.68  & 69.65  & \underline{\textbf{69.58}}  & 69.69  & 69.63  & 69.62  & 69.60  \\  3 & \cite{rice2020overfitting} & 88.67 & 68.64  & 68.60  & \underline{68.55}  & 68.64  & 68.58  & 68.56  & \textbf{68.53}  \\  4 & \cite{rony2019decoupling} & 89.05 & 66.60  & \underline{66.58}  & 66.63  & 66.59  & \underline{66.58}  & \underline{66.58}  & \textbf{66.57}  \\  5 & \cite{Ding2020MMA} & 88.02 & 66.22  & \underline{66.21}  & 66.29  & 66.22  & \underline{66.21}  & \underline{66.21}  & \textbf{66.19}  \\ \multicolumn{10}{c}{}\\ \multicolumn{10}{l}{\textbf{ImageNet} - $\epsilon=3$} \\ \hline 1 & \cite{robustness} & 55.3 & 31.9  & 31.6  & 31.6  & 31.9  & \underline{\textbf{31.5}}  & 31.6  & \textbf{31.5}
	\end{tabular} \end{table*}

\begin{table*}[p]
	\caption{\textbf{Comparison PGD vs \apgd{} on the CW loss in the $l_2$-threat model.} We run PGD and PGD with Momentum (the same used for \apgd{}) on the models tested in Sec. \ref{sec:exps}, with three different step sizes: $\epsilon/10$, $\epsilon/4$ and $2\epsilon$.
		Similarly to \apgd{}, we use 100 iterations and 5 restarts for PGD, and report the resulting robust accuracy on the whole test set (1000 images for ImageNet).
		For each model we boldface the best attack and underline the best version of PGD (i.e. we exclude \apgd{}).}\vspace{2mm}\label{tab:app_pgd_vs_apgd_cw_loss_l2}
	\centering \small \begin{tabular}{r l R{10mm}|| *{3}{R{10mm}}|*{3}{R{10mm}} || *{1}{R{10mm}}} 
		& & & \multicolumn{3}{c|}{PGD} & \multicolumn{3}{c||}{PGD with Momentum} & \\
		\cline{4-9}
		\# & paper & clean & $\epsilon/10$ & $\epsilon/4$ & $2\epsilon$ & $\epsilon/10$ & $\epsilon/4$ & $2\epsilon$ & \apgd\\ \hline
		\multicolumn{10}{c}{}\\ \multicolumn{10}{l}{\textbf{CIFAR-10} - $\epsilon=0.5$}\\
		\hline		
		1 & \cite{augustin2020adversarial} & 91.08 & 74.57  & 74.54  & 74.55  & 74.56  & 74.55  & \underline{74.53}  & \textbf{74.49}  \\  2 & \cite{robustness} & 90.83 & 70.08  & 70.07  & \underline{\textbf{70.06}}  & 70.08  & 70.07  & \underline{\textbf{70.06}}  & \textbf{70.06}  \\  3 & \cite{rice2020overfitting} & 88.67 & 68.73  & \underline{\textbf{68.71}}  & 68.72  & 68.73  & 68.72  & \underline{\textbf{68.71}}  & \textbf{68.71}  \\  4 & \cite{rony2019decoupling} & 89.05 & 67.00  & 67.00  & 67.03  & 67.00  & 67.00  & \underline{\textbf{66.99}}  & \textbf{66.99}  \\  5 & \cite{Ding2020MMA} & 88.02 & 66.55  & 66.54  & 66.64  & 66.55  & 66.54  & \underline{\textbf{66.53}}  & 66.54  \\ \multicolumn{10}{c}{}\\ \multicolumn{10}{l}{\textbf{ImageNet} - $\epsilon=3$} \\ \hline 1 & \cite{robustness} & 55.3 & \underline{\textbf{30.7}}  & \underline{\textbf{30.7}}  & 30.8  & 30.8  & \underline{\textbf{30.7}}  & \underline{\textbf{30.7}}  & \textbf{30.7} \end{tabular}
\end{table*}

\begin{table*}[p]
	\caption{\textbf{Comparison PGD vs \apgd{} on the DLR loss in the $l_2$-threat model.} We run PGD and PGD with Momentum (the same used for \apgd{}) on the models tested in Sec. \ref{sec:exps}, with three different step sizes: $\epsilon/10$, $\epsilon/4$ and $2\epsilon$.
		Similarly to \apgd{}, we use 100 iterations and 5 restarts for PGD, and report the resulting robust accuracy on the whole test set (1000 images for ImageNet).
		For each model we boldface the best attack and underline the best version of PGD (i.e. we exclude \apgd{}).}\vspace{2mm}\label{tab:app_pgd_vs_apgd_new_loss_l2}
	\centering \small \begin{tabular}{r l R{10mm}|| *{3}{R{10mm}}|*{3}{R{10mm}} || *{1}{R{10mm}}} 
		& & & \multicolumn{3}{c|}{PGD} & \multicolumn{3}{c||}{PGD with Momentum} & \\
		\cline{4-9}
		\# & paper & clean & $\epsilon/10$ & $\epsilon/4$ & $2\epsilon$ & $\epsilon/10$ & $\epsilon/4$ & $2\epsilon$ & \apgd\\ \hline
		\multicolumn{10}{c}{}\\ \multicolumn{10}{l}{\textbf{CIFAR-10} - $\epsilon=0.5$}\\
		\hline
		1 &\cite{augustin2020adversarial} & 91.08 & 75.06  & 75.00  & \underline{74.99}  & 75.06  & 75.00  & \underline{74.99}  & \textbf{74.94}  \\  2 & \cite{robustness} & 90.83 & 70.20  & 70.19  & \underline{\textbf{70.17}}  & 70.20  & 70.19  & \underline{\textbf{70.17}}  & 70.20  \\  3 & \cite{rice2020overfitting} & 88.67 & 69.02  & 69.01  & 68.96  & 69.02  & 69.02  & \underline{\textbf{68.94}}  & 68.95  \\  4 & \cite{rony2019decoupling} & 89.05 & 67.03  & \underline{\textbf{67.02}}  & 67.06  & 67.03  & \underline{\textbf{67.02}}  & \underline{\textbf{67.02}}  & \textbf{67.02}  \\  5 & \cite{Ding2020MMA} & 88.02 & 66.61  & 66.56  & 66.64  & 66.61  & 66.55  & \underline{66.54}  & \textbf{66.53}  \\ \multicolumn{10}{c}{}\\ \multicolumn{10}{l}{\textbf{ImageNet} - $\epsilon=3$} \\ \hline 1 & \cite{robustness} & 55.3 & \underline{\textbf{30.9}}  & \underline{\textbf{30.9}}  & \underline{\textbf{30.9}}  & \underline{\textbf{30.9}}  & 31.1  & \underline{\textbf{30.9}}  & \textbf{30.9}
\end{tabular}\end{table*}

\comment{
\begin{table*}[p]
	\caption{\textbf{Comparison PGD vs \apgd{} on the CW loss in the $l_2$-threat model.} We run PGD and PGD with Momentum (the same used for \apgd{}) on the models tested in Sec. \ref{sec:exps}, with three different step sizes: $\epsilon/10$, $\epsilon/4$ and $2\epsilon$.
		Similarly to \apgd{}, we use 100 iterations and 5 restarts for PGD, and report the resulting robust accuracy on the whole test set (1000 images for ImageNet).
		For each model we boldface the best attack and underline the best version of PGD (i.e. we exclude \apgd{}).}\vspace{2mm}
	\centering \small \begin{tabular}{r l R{10mm}|| *{3}{R{10mm}}|*{3}{R{10mm}} || *{1}{R{10mm}}} 
		& & & \multicolumn{3}{c|}{PGD} & \multicolumn{3}{c||}{PGD with Momentum} & \\
		\cline{4-9}
		\# & paper & clean & $\epsilon/10$ & $\epsilon/4$ & $2\epsilon$ & $\epsilon/10$ & $\epsilon/4$ & $2\epsilon$ & \apgd\\ \hline
		\multicolumn{10}{c}{}\\ \multicolumn{10}{l}{\textbf{CIFAR-10} - $\epsilon=8/255$}\\
		\hline		
		1 & \cite{CarEtAl19} & 89.69 & 60.73  & 60.72  & 62.20  & 60.73  & \underline{60.69}  & 61.02  & \textbf{60.48}  \\  2 & \cite{AlaEtAl19} & 86.46 & 61.78  & 61.85  & 63.23  & 61.69  & \underline{61.61}  & 62.10  & \textbf{61.13}  \\  3 & \cite{pmlr-v97-hendrycks19a} & 87.11 & 56.44  & 56.42  & 57.15  & 56.45  & 56.41  & \underline{56.39}  & \textbf{56.20}  \\  4 & \cite{KumEtAl19} & 87.80 & 51.13  & 51.07  & 52.67  & 51.14  & \underline{51.03}  & 51.22  & \textbf{50.88}  \\  5 & \cite{MaoEtAl19} & 86.21 & 49.88  & \underline{49.84}  & 51.18  & 49.88  & \underline{49.84}  & 50.02  & \textbf{49.71}  \\  6 & \cite{ZhaEtAl19-yopo} & 87.20 & 47.13  & 47.09  & 48.27  & 47.11  & \underline{47.06}  & 47.08  & \textbf{46.89}  \\  7 & \cite{MadEtAl2018} & 87.14 & 45.98  & 45.82  & 47.47  & 45.95  & \underline{45.77}  & 46.11  & \textbf{45.67}  \\  8 & \cite{pang2020rethinking} & 80.89 & 44.40  & 44.41  & 45.83  & 44.41  & \underline{44.36}  & 44.66  & \textbf{44.11}  \\  9 & \cite{WonEtAl20} & 83.34 & 46.04  & 45.94  & 47.72  & 46.03  & \underline{45.93}  & 46.26  & \textbf{45.66}  \\  10 & \cite{ShaEtAl19} & 86.11 & 44.89  & 45.93  & 49.27  & \underline{44.16}  & 45.10  & 47.43  & \textbf{43.90}  \\  11 & \cite{ZhaWan19} & 89.98 & 55.98  & 55.30  & \underline{\textbf{48.59}}  & 55.99  & 55.27  & 50.91  & 51.65  \\  12 & \cite{Moosavi-Dezfooli-2019-CVPR} & 83.11 & 40.21  & 40.18  & 40.93  & 40.20  & \underline{40.16}  & 40.22  & \textbf{40.10}  \\  13 & \cite{zhang2020adversarial} & 90.25 & 67.00  & 66.61  & \underline{\textbf{57.87}}  & 67.02  & 66.67  & 64.19  & 64.46  \\  14 & \cite{kim2020sensible} & 91.51 & 56.85  & 55.90  & \underline{\textbf{50.26}}  & 56.94  & 55.92  & 52.54  & 53.04  \\  15 & \cite{Jang-2019-ICCV} & 78.91 & 36.85  & \underline{36.81}  & 39.32  & 36.85  & 36.83  & 37.30  & \textbf{36.52}  \\  16 & \cite{Moosavi-Dezfooli-2019-CVPR} & 80.41 & 35.38  & 35.37  & 36.86  & 35.38  & \underline{35.35}  & 35.48  & \textbf{35.29}  \\  17 & \cite{Wang-2019-ICCV} & 92.80 & 59.93  & 58.42  & \underline{\textbf{52.56}}  & 60.01  & 58.46  & 54.51  & 54.77  \\  18 & \cite{Wang-2019-ICCV} & 92.82 & 66.49  & 65.86  & \underline{\textbf{56.71}}  & 66.60  & 65.76  & 63.53  & 63.80  \\  19 & \cite{Mustafa-2019-ICCV} & 89.16 & 18.27  & 14.75  & \underline{\textbf{4.42}}  & 18.44  & 14.66  & 5.46  & 4.73  \\  20 & \cite{chan2020jacobian} & 93.79 & 1.32  & \underline{1.28}  & 5.51  & 1.30  & 1.29  & 1.29  & \textbf{1.23}  \\  21 & \cite{pang2020rethinking} & 93.52 & 1.88  & 0.54  & 1.42  & 1.59  & 0.53  & \underline{0.47}  & \textbf{0.11}  \\ \multicolumn{10}{c}{}\\ \multicolumn{10}{l}{\textbf{CIFAR-10} - $\epsilon=0.031$} \\ \hline 1 & \cite{ZhaEtAl2019} & 84.92 & 54.05  & 54.06  & 55.16  & 54.08  & \underline{54.03}  & 54.22  & \textbf{53.90}  \\  2 & \cite{AtzEtAl19} & 81.30 & 40.37  & 40.28  & 42.30  & 40.36  & \underline{40.21}  & 40.56  & \textbf{40.05}  \\  3 & \cite{xiao2020enhancing} & 79.28 & 35.77  & 35.53  & \underline{34.04}  & 36.14  & 35.31  & 35.76  & \textbf{32.09}  \\ \multicolumn{10}{c}{}\\ \multicolumn{10}{l}{\textbf{CIFAR-10} - $\epsilon=4/255$} \\ \hline 1 & \cite{song2018improving} & 84.81 & 57.54  & 57.53  & 57.78  & 57.52  & 57.53  & \underline{57.51}  & \textbf{57.49}  \\ \multicolumn{10}{c}{}\\ \multicolumn{10}{l}{\textbf{CIFAR-10} - $\epsilon=0.02$} \\ \hline 1 & \cite{pmlr-v97-pang19a} & 91.22 & 7.12  & 6.09  & 16.15  & 6.87  & \underline{5.57}  & 9.49  & \textbf{3.82}  \\  2 & \cite{pmlr-v97-pang19a} & 93.44 & 0.47  & 0.26  & 3.13  & 0.42  & \underline{0.13}  & 1.22  & \textbf{0.12}  \\ \multicolumn{10}{c}{}\\ \multicolumn{10}{l}{\textbf{MNIST} - $\epsilon=0.3$} \\ \hline 1 & \cite{ZhaEtAl2019} & 99.48 & 94.40  & 94.35  & 95.23  & 94.18  & \underline{94.04}  & 94.33  & \textbf{93.35}  \\  2 & \cite{AtzEtAl19} & 99.35 & 94.43  & 94.69  & 95.98  & 94.03  & \underline{93.74}  & 95.21  & \textbf{92.98}  \\  3 & \cite{MadEtAl2018} & 98.53 & 90.82  & 90.96  & 91.94  & 90.39  & \underline{90.30}  & 91.07  & \textbf{89.39}  \\  4 & \cite{Jang-2019-ICCV} & 98.47 & 93.28  & 93.64  & 95.26  & \underline{92.68}  & 92.71  & 94.49  & \textbf{92.40}  \\  5 & \cite{WonEtAl20} & 98.50 & 87.32  & 87.55  & 90.20  & 86.74  & \underline{86.62}  & 87.25  & \textbf{84.97}  \\  6 & \cite{Taghanaki-2019-CVPR} & 98.86 & \underline{\textbf{0.00}}  & \underline{\textbf{0.00}}  & \underline{\textbf{0.00}}  & \underline{\textbf{0.00}}  & \underline{\textbf{0.00}}  & \underline{\textbf{0.00}}  & \textbf{0.00}
\end{tabular} \end{table*}}

\comment{
\begin{table*}\caption{\textbf{Comparison FAB+Square vs \Combo{}.}}\vspace{2mm}\centering \small
	\begin{tabular}{r l R{10mm}|| *{3}{R{15mm}}} \# & paper & clean & FAB+Square & AA & reduct. \\
		\hline \multicolumn{6}{c}{}\\ \multicolumn{6}{l}{\textbf{CIFAR-10} - $\epsilon=8/255$} \\
	\hline
	1 & \cite{CarEtAl19} & 89.69 & 60.49  & 59.65  & -0.84  \\  2 & \cite{AlaEtAl19} & 86.46 & 58.18  & 56.92  & -1.26  \\  3 & \cite{pmlr-v97-hendrycks19a} & 87.11 & 55.39  & 54.99  & -0.40  \\  4 & \cite{KumEtAl19} & 87.80 & 49.84  & 49.40  & -0.44  \\  5 & \cite{MaoEtAl19} & 86.21 & 48.28  & 47.66  & -0.62  \\  6 & \cite{ZhaEtAl19-yopo} & 87.20 & 45.61  & 45.06  & -0.55  \\  7 & \cite{MadEtAl2018} & 87.14 & 45.31  & 44.29  & -1.02  \\  8 & \cite{pang2020rethinking} & 80.89 & 44.34  & 43.78  & -0.56  \\  9 & \cite{WonEtAl20} & 83.34 & 44.01  & 43.38  & -0.63  \\  10 & \cite{ShaEtAl19} & 86.11 & 42.59  & 41.58  & -1.01  \\  11 & \cite{ZhaWan19} & 89.98 & 39.85  & 38.78  & -1.07  \\  12 & \cite{Moosavi-Dezfooli-2019-CVPR} & 83.11 & 39.03  & 38.67  & -0.36  \\  13 & \cite{zhang2020adversarial} & 90.25 & 40.02  & 38.57  & -1.45  \\  14 & \cite{kim2020sensible} & 91.51 & 37.80  & 36.10  & -1.70  \\  15 & \cite{Jang-2019-ICCV} & 78.91 & 35.48  & 35.09  & -0.39  \\  16 & \cite{Moosavi-Dezfooli-2019-CVPR} & 80.41 & 34.07  & 33.76  & -0.31  \\  17 & \cite{Wang-2019-ICCV} & 92.80 & 33.29  & 30.96  & -2.33  \\  18 & \cite{Wang-2019-ICCV} & 92.82 & 31.62  & 29.07  & -2.55  \\  19 & \cite{Mustafa-2019-ICCV} & 89.16 & 1.09  & 0.55  & -0.54  \\  20 & \cite{chan2020jacobian} & 93.79 & 13.77  & 0.18  & -13.59  \\  21 & \cite{pang2020rethinking} & 93.52 & 0.00  & 0.00  & 0.00  \\ \multicolumn{6}{c}{}\\ \multicolumn{6}{l}{\textbf{CIFAR-10} - $\epsilon=0.031$} \\ \hline 1 & \cite{ZhaEtAl2019} & 84.92 & 53.70  & 53.18  & -0.52  \\  2 & \cite{AtzEtAl19} & 81.30 & 40.78  & 40.61  & -0.17  \\  3 & \cite{xiao2020enhancing} & 79.28 & 20.44  & 17.99  & -2.45  \\ \multicolumn{6}{c}{}\\ \multicolumn{6}{l}{\textbf{CIFAR-10} - $\epsilon=4/255$} \\ \hline 1 & \cite{song2018improving} & 84.81 & 57.29  & 56.64  & -0.65  \\ \multicolumn{6}{c}{}\\ \multicolumn{6}{l}{\textbf{CIFAR-10} - $\epsilon=0.02$} \\ \hline 1 & \cite{pmlr-v97-pang19a} & 91.22 & 4.38  & 2.00  & -2.38  \\  2 & \cite{pmlr-v97-pang19a} & 93.44 & 0.13  & 0.02  & -0.11  \\ \multicolumn{6}{c}{}\\ \multicolumn{6}{l}{\textbf{MNIST} - $\epsilon=0.3$} \\ \hline 1 & \cite{ZhaEtAl2019} & 99.48 & 92.93  & 92.76  & -0.17  \\  2 & \cite{AtzEtAl19} & 99.35 & 90.85  & 90.85  & 0.00  \\  3 & \cite{MadEtAl2018} & 98.53 & 88.53  & 88.43  & -0.10  \\  4 & \cite{Jang-2019-ICCV} & 98.47 & 88.00  & 87.99  & -0.01  \\  5 & \cite{WonEtAl20} & 98.50 & 83.04  & 82.88  & -0.16  \\  6 & \cite{Taghanaki-2019-CVPR} & 98.86 & 0.00  & 0.00  & 0.00
\end{tabular}\end{table*}
}

%
%
%
%

%

%
%

%

%
%


\fi

\end{document}